\documentclass[10pt]{article} 
\usepackage[accepted]{tmlr}

\usepackage{amsmath,amsfonts,bm}

\newcommand{\tabref}[1]{Table~\ref{tbl:#1}}

\newcommand{\figref}[1]{Figure.~\ref{fig:#1}}

\def\eqref#1{equation~\ref{#1}}

\def\1{\bm{1}}

\DeclareMathAlphabet{\mathsfit}{\encodingdefault}{\sfdefault}{m}{sl}
\SetMathAlphabet{\mathsfit}{bold}{\encodingdefault}{\sfdefault}{bx}{n}

\usepackage{hyperref}
\usepackage{url}
\usepackage{graphicx}
\usepackage{subcaption}
\usepackage{booktabs}
\usepackage{bbding}

\PassOptionsToPackage{pagebackref,breaklinks,colorlinks}{hyperref}
\usepackage{amsmath}
\usepackage[utf8]{inputenc}   
\usepackage{textcomp}         
\DeclareUnicodeCharacter{2218}{\ensuremath{\circ}} 

\title{Mixed-View Panorama Synthesis using Geospatially Guided Diffusion}

\author{\name Zhexiao Xiong \email x.zhexiao@wustl.edu \\
      \addr Department of Computer Science $\&$ Engineering, Washington University in St. Louis
      \AND
      \name Xin Xing \email xxing@unomaha.edu \\
      \addr Department of Computer Science, University of Nebraska Omaha
      \AND
      \name Scott Workman \email scott.workman.ai@gmail.com \\
      \addr DZYNE Technologies 
        \addr 
    \AND
      \name Subash Khanal  \email k.subash@wustl.edu \\
      \addr Department of Computer Science $\&$ Engineering, Washington University in St. Louis
      \AND
      \name Nathan Jacobs  \email jacobsn@wustl.edu \\
      \addr Department of Computer Science $\&$ Engineering, Washington University in St. Louis
      }

\def\month{05}  
\def\year{2025} 
\def\openreview{\url{https://openreview.net/forum?id=ylUVRikhTL}} 

\begin{document}

\maketitle

\begin{abstract}
We introduce the task of mixed-view panorama synthesis, where the goal is to synthesize a novel panorama given a small set of input panoramas and a satellite image of the area. This contrasts with previous work which only uses input panoramas (same-view synthesis), or an input satellite image (cross-view synthesis). We argue that the mixed-view setting is the most natural to support panorama synthesis for arbitrary locations worldwide. A critical challenge is that the spatial coverage of panoramas is uneven, with few panoramas available in many regions of the world. We introduce an approach that utilizes diffusion-based modeling and an attention-based architecture for extracting information from all available input imagery. Experimental results demonstrate the effectiveness of our proposed method. In particular, our model can handle scenarios when the available panoramas are sparse or far from the location of the panorama we are attempting to synthesize. The project page is available at \url{https://mixed-view.github.io}.
\end{abstract}

\section{Introduction}

The wide availability of street-level panoramas and their integration into mapping applications has dramatically improved the navigation experience for users. Access to nearby panoramas reduces the difficulty of navigating from a purely overhead map.  However, an inherent problem is that panoramas are expensive to collect and thus are sparsely available and updated infrequently, with many roads having no panoramas. This has motivated the task of cross-view synthesis~\cite{zhai2017predicting,regmi2018cross,9674229}, where street-level panoramas are synthesized directly from satellite imagery. Unfortunately, existing approaches ignore any available nearby street-level panoramas at inference time.

\begin{figure}[t]
    \centering
    \includegraphics[width=\linewidth]{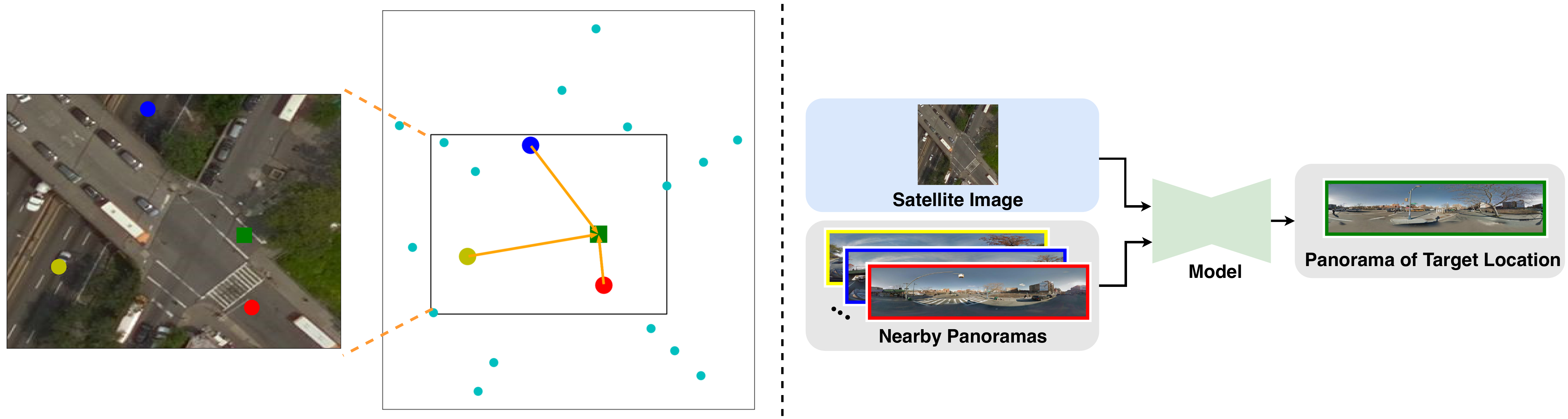}
    \caption{We propose a new task, mixed-view panorama synthesis, in which a satellite image and a set of nearby panoramas (blue, yellow, and green) are used to render a panorama at a novel location (green). Our approach uses diffusion-based modeling and attention to enable flexible, multimodal control.}
    \label{fig:combined_data_vis}
\end{figure}

We propose a new task, mixed-view panorama synthesis (MVPS), which also aims to synthesize a street-level panorama. However, available street-level panoramas are also used to control the synthesis process. \figref{combined_data_vis} gives a visual overview of the proposed mixed-view setting. In contrast to many recent works on novel view synthesis of outdoor scenes (e.g., NeRF-related works~\citep{mildenhall2021nerf, martin2021nerf,xie2023s,tancik2022block} and 3DGS-related works~\citep{kerbl3Dgaussians,zhou2024drivinggaussian,liu2024citygaussian}), which require dense images with accurate camera pose information captured under fairly controlled settings, our input data are sparsely distributed panoramas, with only geo-location and orientation information provided. In particular, the distance, orientation, and availability of nearby street-level panoramas can vary dramatically. This task is in many ways more challenging than the text-to-image generation problem. First, previous text-to-image generation methods only focus on semantic accuracy, while in our task, geometric faithfulness is the primary factor to be considered. Second, nearby panoramas are often not acquired simultaneously, leading to significant challenges with non-stationary objects, lighting variation, and seasonal changes.  Therefore, a method needs to be able to condition the output on images from different viewpoints, both street-level and overhead, using the geometric relationship between them.

Recently, conditional diffusion models~\cite{Rombach_2022_CVPR} have been widely used in the image synthesis task. However, most of them are typically restricted to single-condition inputs, which is unsuitable for our MVPS task that aims to use multiple conditions and their correspondence to synthesize complex outdoor scenes. ControlNet~\cite{zhang2023adding} extends the pretrained Stable Diffusion model to take multi-modal data as inputs, while ControlNet-based approaches still focus primarily on conditional inputs that are spatially aligned with the target image, such as sketches, depth maps, and segmentation maps, which are essentially different forms of representation of the target image.  In addition, recent works on attention-manipulated diffusion models predominantly focus on manipulating the text modality via cross-attention module. For outdoor panorama synthesis, with complex scene layouts, text alone is not sufficient to achieve fine-grained spatial control. Limitations of current diffusion frameworks motivate the need for frameworks that allow multi-conditioned and fine-grained geometry control for outdoor panorama synthesis. To address these challenges, there is a pressing need to develop a framework capable of handling multiple image input controls from diverse modalities, with the ability to guide the model to focus on the regions of the input images with salient geometric correspondence to the target image.  

In this work, we propose a multi-conditioned, end-to-end diffusion framework for combining information from all input imagery to guide the diffusion-based mixed-view synthesis process. A key element of our approach is the integration of geospatial attention~\cite{workman2022revisiting} to guide the controllable diffusion model for MVPS task. Geospatial attention incorporates the semantic content of the input, geometry, and overhead appearance to identify the geo-informative regions of an input panorama relative to the target location. We extend the concept of geospatial attention to include both local-level and global-level attention. The resulting attention maps are fused with the corresponding input images in the latent space of the diffusion model. Finally, the attention-guided features are integrated into the encoders of the corresponding conditional branch in the multi-conditioned ControlNet model, achieving geometric-guided, fine-grained spatial control. 

The key contributions of this work can be summarized as:
\begin{itemize}
    \item We propose the mixed-view panorama synthesis task: using a satellite image and a sparse set of nearby panoramas to synthesize a target panorama at a given location.
    \item We propose a unified multi-conditioned controllable diffusion framework, GeoDiffusion, for the mixed-view panorama synthesis task.
    \item We use geospatial attention as the geometric constraint to associate nearby views with the target view to achieve geometry-guided fine-grained spatial control.
    \item We validate that our proposed method generates high-fidelity panorama images with geometric accuracy on the Brooklyn and Queens benchmark dataset, achieving state-of-the-art performance while having more flexibility compared with prior cross-view synthesis works.
\end{itemize}

\section{Related Work}
\subsection{Cross-view Synthesis}
 
Given a satellite image, the cross-view synthesis task aims to predict the street-level panorama. Prior work includes using a learned linear transformation~\cite{zhai2017predicting}, applying conditional GANs~\cite{regmi2018cross, regmi2019cross, tang2019multi,li2021sat2vid}, integrating height estimation as explicit supervision~\cite{9674229}, using a density field representation~\cite{qian2023sat2density} and point-based neural rendering~\cite{li2024sat2scene}. Structure-preserving panorama synthesis methods have also shown to be effective in the related task of cross-view image geolocalization~\cite{regmi2019bridging, toker2021coming, shi2022accurate}. However, such methods typically only consider a satellite image as input and only attempt to synthesize street-level panoramas in the setting where the satellite image is center-aligned with the target location. Our work integrates near and remote modalities (i.e., mixed-view) and can synthesize panoramas at arbitrary locations in the given satellite image region. 

\subsection{Image-to-Image Translation}

The goal of image-to-image translation is to learn a mapping between an input image and a target image~\cite{isola2017image}. Traditional methods are based on generative adversarial networks~\cite{park2019semantic,wang2018high,liu2017unsupervised,zhu2017unpaired}. With the development of vision transformers, several works have successfully applied transformers to this problem~\cite{esser2021taming,Chang_2022_CVPR}. More recently, numerous methods have leveraged diffusion model to perform image translation~\cite{dhariwal2021diffusion, Rombach_2022_CVPR,xue2023freestyle,parmar2023zero,wang2022pretraining}. However, these methods are unable to handle cases in which direct conditions are unknown and only implicit geometric relationships are available. In contrast, our model is capable of handling multiple indirect image controls and utilizing the geometric relationships both within and across different input modalities.

\subsection{Conditional Diffusion Models}

Conditional diffusion models enable controllable image synthesis and editing. Most recent work in this domain focuses on the text-to-image synthesis problem, such as GLIDE~\cite{nichol2021glide}, DALL-E2~\cite{ramesh2022hierarchical} and Stable Diffusion~\cite{Rombach_2022_CVPR}. These methods require large training datasets containing many image-text pairs to generate high-quality images. However, generating complex scenes is challenging with only text information. Recent works have proposed incorporating local, spatial conditioning, such as segmentation maps~\cite{mou2023t2i} or layouts~\cite{li2023gligen,zheng2023layoutdiffusion,qu2023layoutllm}, to overcome these challenges and achieve precise spatial control. ControlNet~\cite{zhang2023adding} and similar methods~\cite{qin2023unicontrol,zhao2023uni} use zero-convolution layers to incorporate task-specific conditioning into pretrained image diffusion models, significantly reducing the computational cost and sample complexity while still generating high-quality images. We adopt this approach to build a multi-conditioned, end-to-end geospatial attention-guided diffusion framework.

\subsection{Attention}

Attention mechanisms have shown benefits in a variety of visual tasks. Commonly used attention mechanisms include channel-wise attention~\cite{hu2018squeeze, woo2018cbam}, spatial attention~\cite{mnih2014recurrent, jaderberg2015spatial, hu2018gather, Wang_2018_CVPR}, spatial-temporal attention~\cite{meng2019interpretable, song2017end}, and branch attention~\cite{Fukui_2019_CVPR}. Self-attention~\cite{vaswani2017attention} and cross-attention~\cite{Chen_2021_ICCV} are widely applied in vision transformers~\cite{dosovitskiy2020image, carion2020end, xie2021segformer}. Recently latent diffusion models~\cite{Rombach_2022_CVPR,wu2023harnessing,xue2023freestyle}, also regard cross-attention as an effective way to allow multi-modal training for class-conditional, text-to-image, and spatially conditioned tasks. Besides cross-attention, geometry has been used to inform the learning of attention, such as using epipolar attention~\cite{he2020epipolar} in novel view synthesis~\cite{tseng2023consistent}. Workman \textit{et al.}~\cite{workman2022revisiting} proposed the concept of geospatial attention, a geometry-aware attention mechanism that explicitly considers the geospatial relationship between the pixels in a ground-level image and a geographic location. We extend this concept to our mixed-view panorama synthesis task to identify geo-informative regions across mixed viewpoints, allowing our diffusion model to be `geometry aware.'

\section{Preliminaries}






\subsection{Stable Diffusion Architecture}

Stable Diffusion~\cite{Rombach_2022_CVPR} is a generative modeling approach via a learned diffusion process, by applying the diffusion operation in the latent space. It uses a UNet-like structure as its denoising model, which consists of an encoder, a middle block, and a decoder. Both the encoder and the decoder are made up of 12 blocks, and the full model contains 25 blocks. The outputs of the encoder are added to the 12 skip-connections and 1 middle block of the UNet. The input for the $i$-th block in the decoder is represented as:
\begin{equation}
    \left\{\begin{array}{ll}
        \operatorname{concat}\left(m, f_j\right) & \text { where } i=1, j=13-i \\
        \operatorname{concat}\left(g_{i-1}, f_j\right) & \text { where } 2 \leq i \leq 12, j=13-i
    \end{array}\right.
\end{equation}
where $m$ denotes the output of the middle block and $f_i$ and $g_i$ denote the output of the $i$-th block in the encoder and decoder, respectively.

\subsection{Mixed-View Panorama Synthesis}

We introduce the task of mixed-view panorama synthesis (MVPS). Given a satellite image $S_1$, and a set of sparsely distributed nearby street-level panoramas $(P_1, P_2,\cdots, P_n)$ with known geolocations $(l_1, l_2,\cdots, l_n)$, we aim to synthesize a panorama $P_t$ in the region of $S_1$ at a target location $l_t$. To generate the target panorama $P_t$ with precise layout distribution, the synthesis process should utilize: (1) the semantic information from $P_1, P_2,\cdots, P_n$ and $S_1$, (2) the geometric relationships between street-level ${P_i}\to{P_t}, i\in(1,2,\cdots,n)$ and across satellite \& street-level $S_1\to{P_t}$. With only location information provided, how to utilize the implicit geometric relationships is the key challenge of this task.

\section{Geospatial Attention-Guided Diffusion Model}

\begin{figure}[t!]
    \centering
    \includegraphics[width=0.98\linewidth]{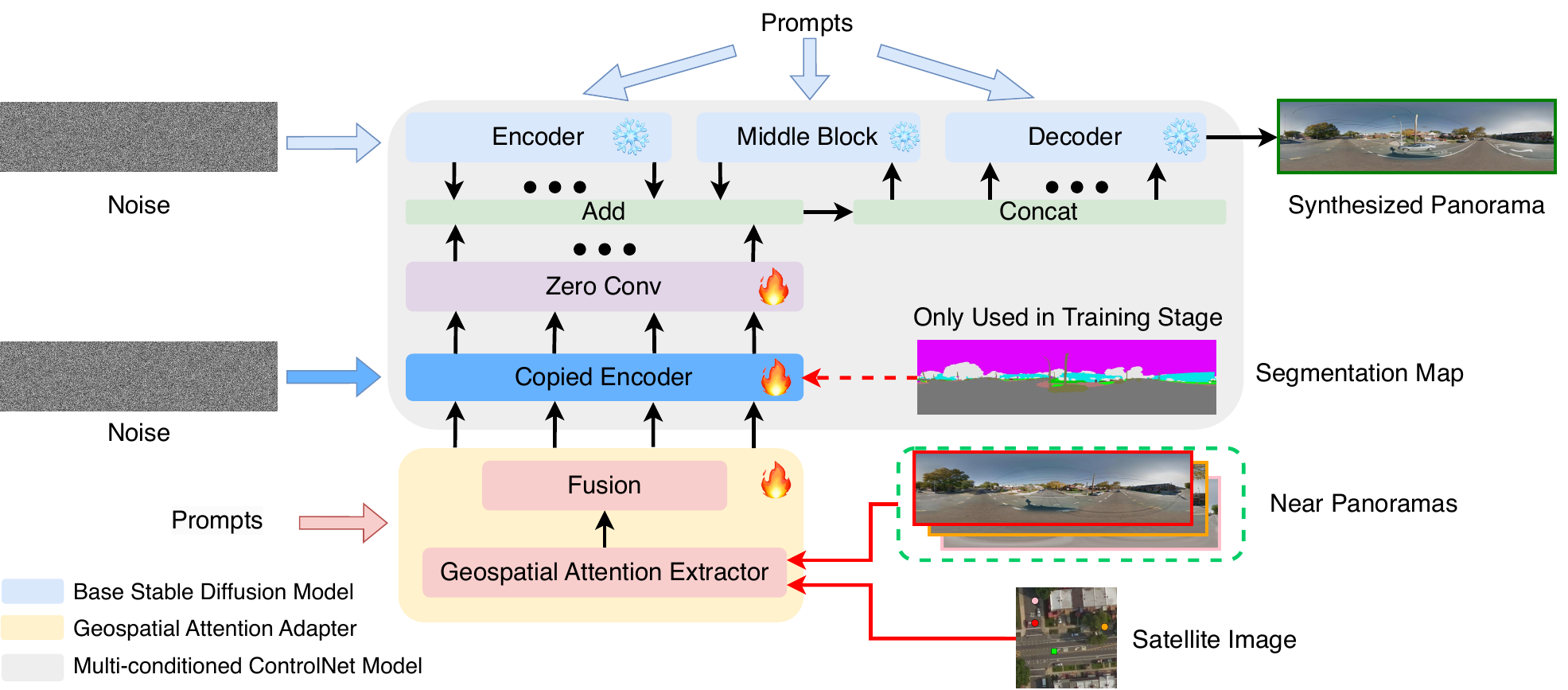}
    \caption{The overall framework of our model. For each branch of the near panoramas in the multi-conditioned ControlNet diffusion model, the near panoramas and the satellite image are passed through the geospatial attention adapter and the copied encoder, and the extracted features are injected into the Stable Diffusion encoder.}
    \label{fig:framework}
\end{figure}

We propose a novel multi-conditioned, geospatial attention-guided diffusion model, GeoDiffusion, which combines information from a satellite image and nearby street-level panoramas to synthesize a target panorama. \figref{framework} shows an overview of our framework, which consists of two main modules: (1) a novel geospatial attention adapter that combines information from the input imagery and (2) a multi-conditioned diffusion model, based on ControlNet~\cite{zhang2023adding}, that synthesizes the target panorama.


\subsection{Geospatial Attention Adapter}
\label{sec:geoattention}

We propose an adapter that uses geospatial attention~\cite{workman2022revisiting} to fuse image features that are extracted using CNN-based encoders. The geospatial attention adapter has two components: (1) local attention for assigning weight to panorama features and (2) global attention for assigning weight to satellite features. These attention maps are used to fuse features from the CNN-derived feature maps, which subsequently control the diffusion model.


\begin{figure*}[t!]
    \centering
    \includegraphics[width=\linewidth]{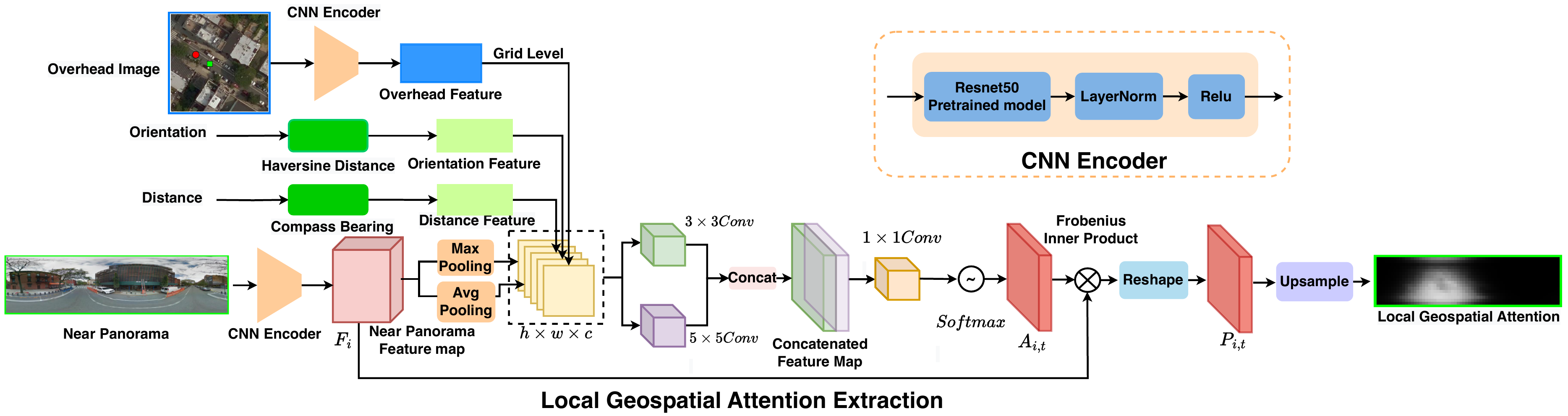}
    \caption{The local geospatial attention extraction module, which provides attention for near panoramas. The red dot represents the target location, and the green dot represents the source panorama location in the overhead image region.}
    \label{fig:attention_extraction}
\end{figure*}
\paragraph{\textbf{Extracting Local Attention}}


To identify geoinformative regions in a nearby street-level panorama, we build a local geospatial attention module, shown visually in \figref{attention_extraction}. In our setting, we assume images are fully calibrated. Given a panorama at location $l_i$ and the target location $l_t$, for MVPS, we apply geospatial attention~\cite{workman2022revisiting} based on the relative location, the target-relative orientation of each pixel ray, the semantic content of the input, and the overhead appearance, and get a spatial attention map, $A_{i,t} \in R^{H \times W}$. Based on $A_{i,t}$ and the feature map of the near panorama $F_i\in R^{H \times W \times C}$, we compute an attention-weighted feature descriptor $P_{i, t}$ as \eqref{attention1}:
\begin{equation}
    P_{i, t}=reshape(\left\langle\mathbf{f}^c, A_{i, t}\right\rangle_F)
    \label{attention1}
\end{equation}
where $\mathbf{f}^c \in R^{H \times W}$ represents the input feature map of the $c$-th channel of $F_i$, $\langle.,.\rangle_F$ denotes the Frobenius inner product of the two inputs, and $P_{i, t}$ represents the feature output, which represents the information from the input feature map $F_i$ that is relevant to the target location $l_t$. We detail the extraction of local-level geospatial attention in the supplementary material.



\paragraph{\textbf{Extracting Global Attention}}

\begin{figure}[t!]
    \centering
    \includegraphics[width=0.55\linewidth]{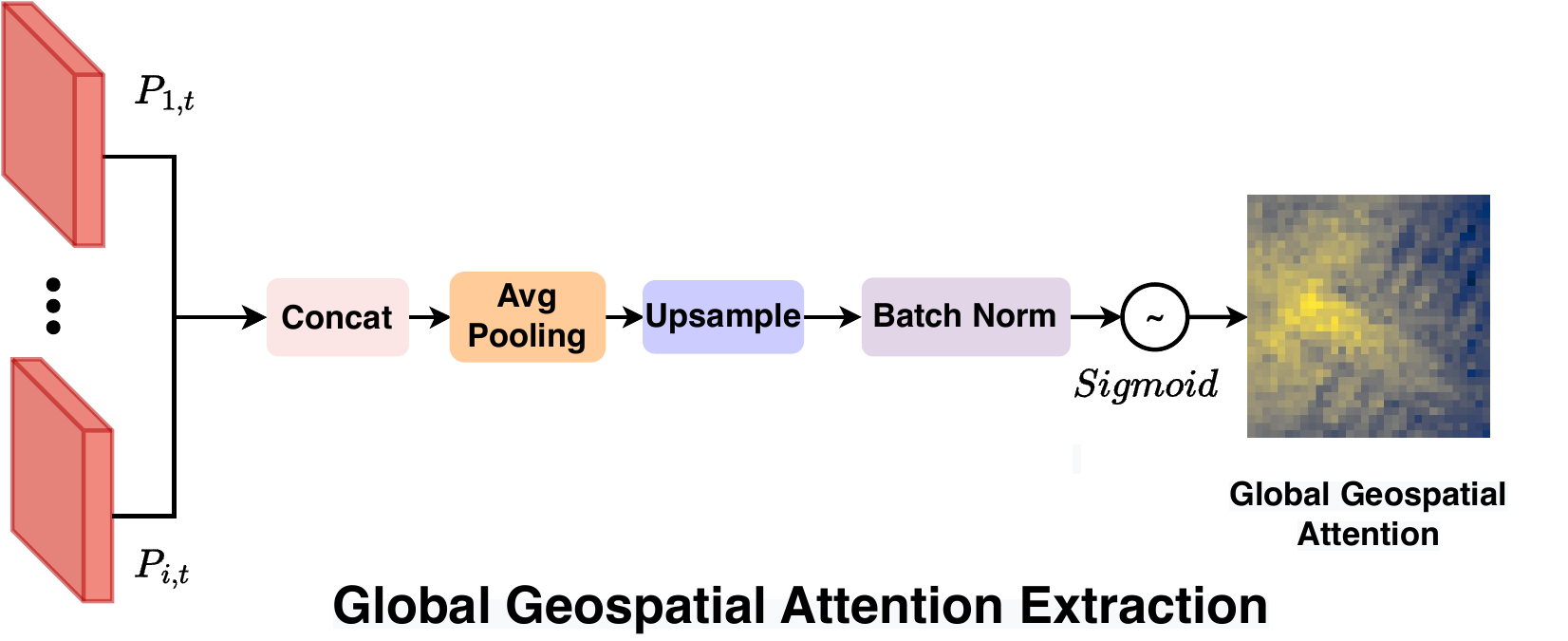}
    \caption{Global geospatial attention extraction module, which provides attention for satellite images.}
    \label{fig:global_attention_extraction}
\end{figure}

One of the key challenges in MVPS is handling the nonuniform spatial distribution of the input panoramas. To address this, we extract the spatial distributions of the panorama features in the given region and propose global geospatial attention, which is an attention map that is used to guide which features should be more heavily weighted in the feature space of satellite image.
See \figref{global_attention_extraction} for an overview. Given a target location $l_t$, this module operates on the attention-weighted features for the input street-level panoramas computed using local-level geospatial attention $ P_{1,t},P_{2,t},\cdots, P_{i,t} $, where $i$ is the number of panoramas. We aggregate those features using concatenation and average pooling, and get a $32\times32$ weight feature grid. We then upsample this dense grid to $256\times256$ and pass it through a batch normalization layer and a sigmoid layer to obtain the final global-level geospatial attention. The resulting attention map is the global-level feature distribution in the overhead-view depending on the relative location of the target panorama and the input panoramas.

\begin{figure*}[t!]
    \centering
    \includegraphics[width=
    \linewidth]{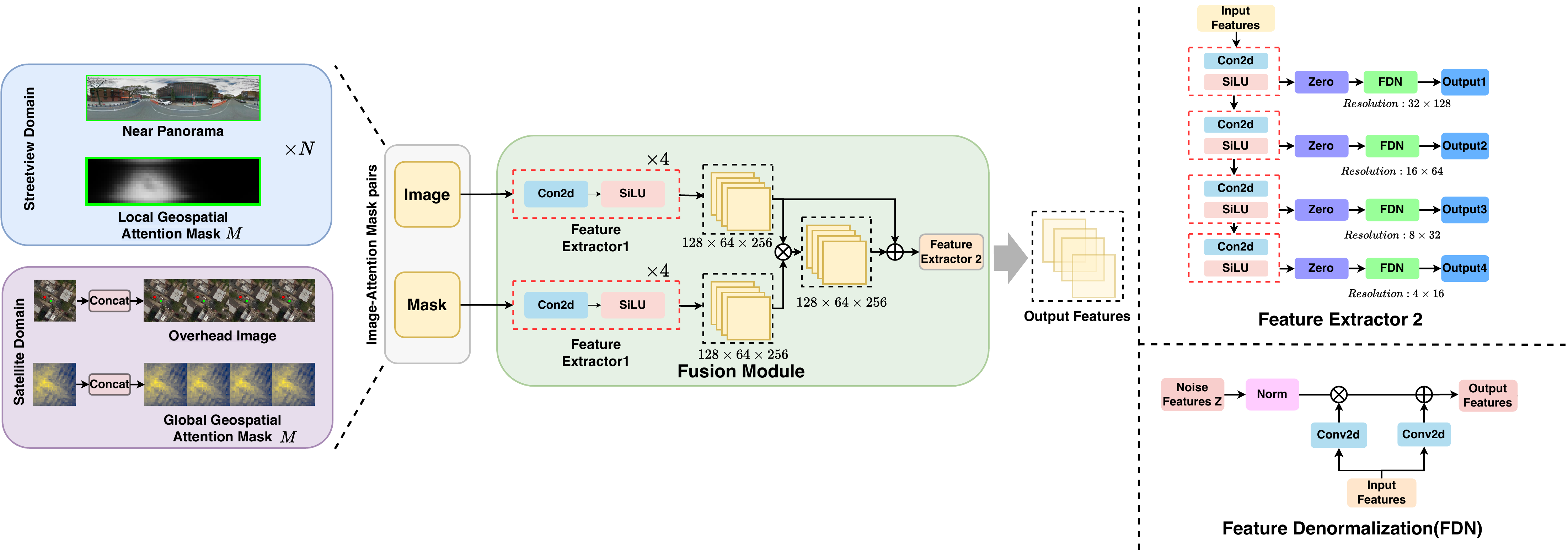}
    \caption{The geospatial attention fusion module. We use the same architecture to add attention to near panoramas and satellite image respectively in the latent space, and inject the features into the copied encoder of the diffusion model. }
    \label{fig:fusion}
\end{figure*}

\paragraph{\textbf{Fusion}}

After extracting the attention mask from both street-level and satellite views, we fuse them with the corresponding street-level panoramas and the satellite image. The fusion module is shown in the left part of \figref{fusion}. For the street-level panoramas, we use a feature extractor composed of stacked convolutions to extract the attention mask and the near panoramas to the latent space. Both of their shapes are $B\times 128\times 64\times 256$ and we fuse them using a Hadamard product. We also add skip connections between the masked near panorama feature map and the original near panorama feature map, which is shown as:
\begin{equation}
    H_{i, c}(x)=\left(1+M_{i, c}(x)\right) \odot F_{i, c}(x)
\end{equation}
where $M(x)$ is the latent space geospatial attention mask ranging from $[0,1]$, $i$ ranges over all spatial positions, and $c \in\{1, \cdots, C\}$ is the index of the channel. When $M(x)$ approximates 0, the output feature $H(x)$ will approximate original latent features $F(x)$.
For the satellite image, based on the extracted satellite attention, we use the same method as nearby street-level panoramas to get the attention-guided feature map. Finally, the output features are passed through a multi-scale feature extractor to get the output features. 

We show the detailed structure of the multi-scale feature extractor in the right part of \figref{fusion}. In the multi-scale feature extractor, for each resolution, we first pass the feature through a zero-convolution layer, which progressively starts influencing the generation with the attention-guided features. Then we adopt Feature Denormalization (FDN)~\cite{park2019semantic} as injection module, which uses the condition features for GeoDiffusion to rectify its normalized input noise features as:
\begin{equation}
    \begin{aligned}
    F D N_r(Z_r, c_l)=norm(Z_r) \cdot(1+conv_\gamma(zero(h_r(c_l)))) 
    +conv_\beta(z e r o(h_r(c_l))),
    \end{aligned}
\end{equation}
where $Z_r$ is the noise features at resolution $r$, $c_r$ represents the concatenated conditions, $h_r$ represents the output of the feature extractor $H$ at resolution $r$, and $conv_{\gamma}$ and $conv_{\beta}$ refer to the learnable convolutional layers that convert condition features into spatial-sensitive scales and shift modulation coefficients respectively. The final extracted features are concatenated and injected into the respective copied encoders. We select the first block of each resolution in the copied encoder for feature injection, corresponding to the $1,4,7,10$ layers of the copied encoder respectively. Leveraging the effectiveness of geospatial attention, the multi-conditioned diffusion network can generate more geometrically accurate panoramas.

\subsection{Multi-Conditioned Diffusion Model}

We introduce our approach based on the conditioning features extracted from the overhead imagery and nearby panoramas. Inspired by ControlNet\cite{zhang2023adding}, we construct a multi-conditioned diffusion model for MVPS (\figref{framework}). We start from a pretrained Stable Diffusion~\cite{Rombach_2022_CVPR} model and duplicate the structures and weights of the encoder and middle block for each condition $i$, which we refer to as $F'_i$ and $M'_i$ respectively. The conditions are concatenated in the channel dimension, passed through the geospatial attention adapter, and then the attention-guided features for each condition are injected into the corresponding copied encoder, with noise added according to the time embedding. The encoded features are passed through a zero-convolution layer and incorporated into the main branch of the Stable Diffusion architecture. 
During the decoding process, we keep all other elements unchanged while modifying the input of the $i$-th block of the decoder as:
\begin{equation}
    \left\{
    \begin{aligned}
        &concat\left(m+m^{\prime},f_j+zero\left(f_j^{\prime}\right)\right) 
        \text { where } i=1,  j=13-i \\
        &concat\left(g_{i-1}, f_j+zero\left(f_j^{\prime}\right)\right) 
        \text { where } 2 \leq i \leq 12, j=13-i
    \end{aligned}
    \right.
\end{equation}
in which $m$ denotes the output of the middle block, $f_i$ and $g_i$ denote the output of the $i$-th block in the encoder and decoder of UNet respectively, and $zero$ represents a zero convolutional layer. Following ControlNet~\cite{zhang2023adding}, the weights of zero convolutional layer is set to increase from zero to gradually add control information into the main Stable Diffusion model. For the text input, as we do not rely on text embedding to achieve content manipulation, we use ``A high-resolution street-view panorama'' as the default prompt.

\section{Experiments}

We evaluate our approach for mixed-view panorama synthesis quantitatively and qualitatively through various experiments. Results show that our approach, which can take advantage of nearby street-level panoramas, significantly improves results compared to existing cross-view methods, and shows more flexibility in synthesizing the panorama in arbitrary locations.

\paragraph{\textbf{Dataset}} 
We train and evaluate our methods using the Brooklyn and Queens dataset~\cite{workman2017unified}. This dataset contains non-overlapping satellite images (approx. 30 cm resolution) and street-level panoramas from New York City collected from Google Street View. It is composed of two subsets collected from Brooklyn and Queens respectively. The Brooklyn subset contains 43,605 satellite images and 139,327 panoramas. The Queens subset, which we use solely for cross-domain evaluation, contains 10,044 satellite images and 38,630 panoramas. For evaluation on the Brooklyn subset, we use the original train/test split, resulting in 38,744 images for training, 500 images for validation, and 4361 images for testing. For cross-domain evaluation on Queens, we randomly select 1000 images from the Queens subset and report the performance on the selected images. During training, when computing geospatial attention, we set the number of nearby street-level panoramas considered, $N$, to 20 for each satellite image; In the conditional diffusion module, for each pair of data, we use the 2 closest panoramas to the target location as conditions. During the inference stage,  the geospatial attention module is frozen, so only 2 nearby panoramas are in need.

\vspace{-0.3cm}
\paragraph{\textbf{Metrics}} 
For evaluating the performance of our approach, we consider two classes of metrics. The first class is low-level metrics which evaluate the pixel-wise similarity between two images: peak signal-to-noise ratio (PSNR), structural similarity index (SSIM), root-mean-square error (RMSE), and sharpness difference (SD). The second class is high-level metrics which evaluate image-level differences between two images. For perceptual similarity (LPIPS)~\cite{zhang2018unreasonable}, we use a pretrained AlexNet~\cite{krizhevsky2012imagenet} and denote it as $P_{alex}$. We also adopt Fréchet inception distance (FID)~\cite{heusel2017gans}, a common metric for measuring the realism and diversity of images produced by generative models.

\begin{figure*}[t!]
    \centering
    \captionsetup[subfigure]{font=small} 
    \begin{subfigure}[b]{0.088\linewidth}
        \includegraphics[width=\linewidth]{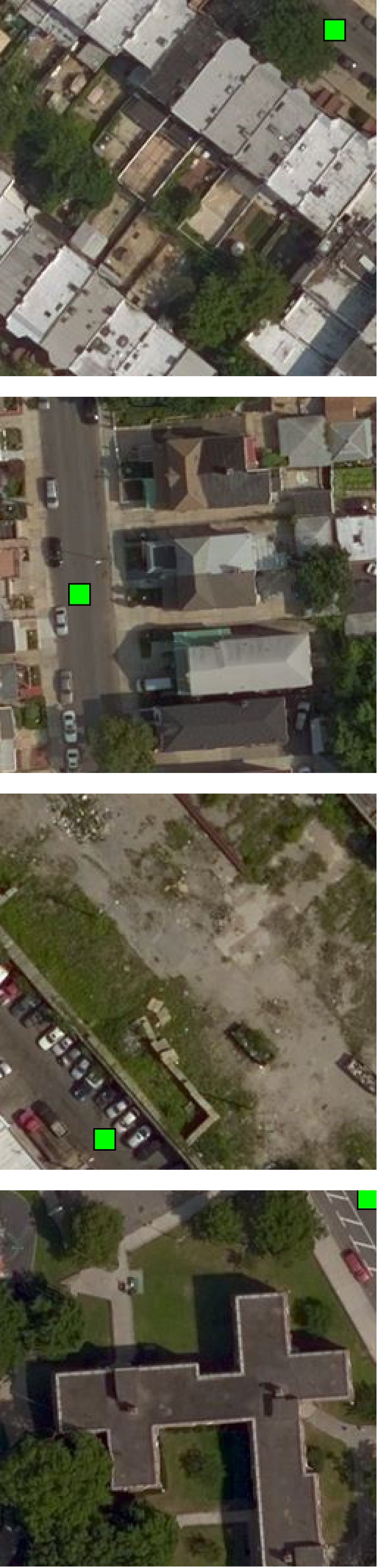}
        \caption{Sat}
        \label{fig:1a}
    \end{subfigure}
    %
    \begin{subfigure}[b]{0.176\linewidth}
        \includegraphics[width=\linewidth]{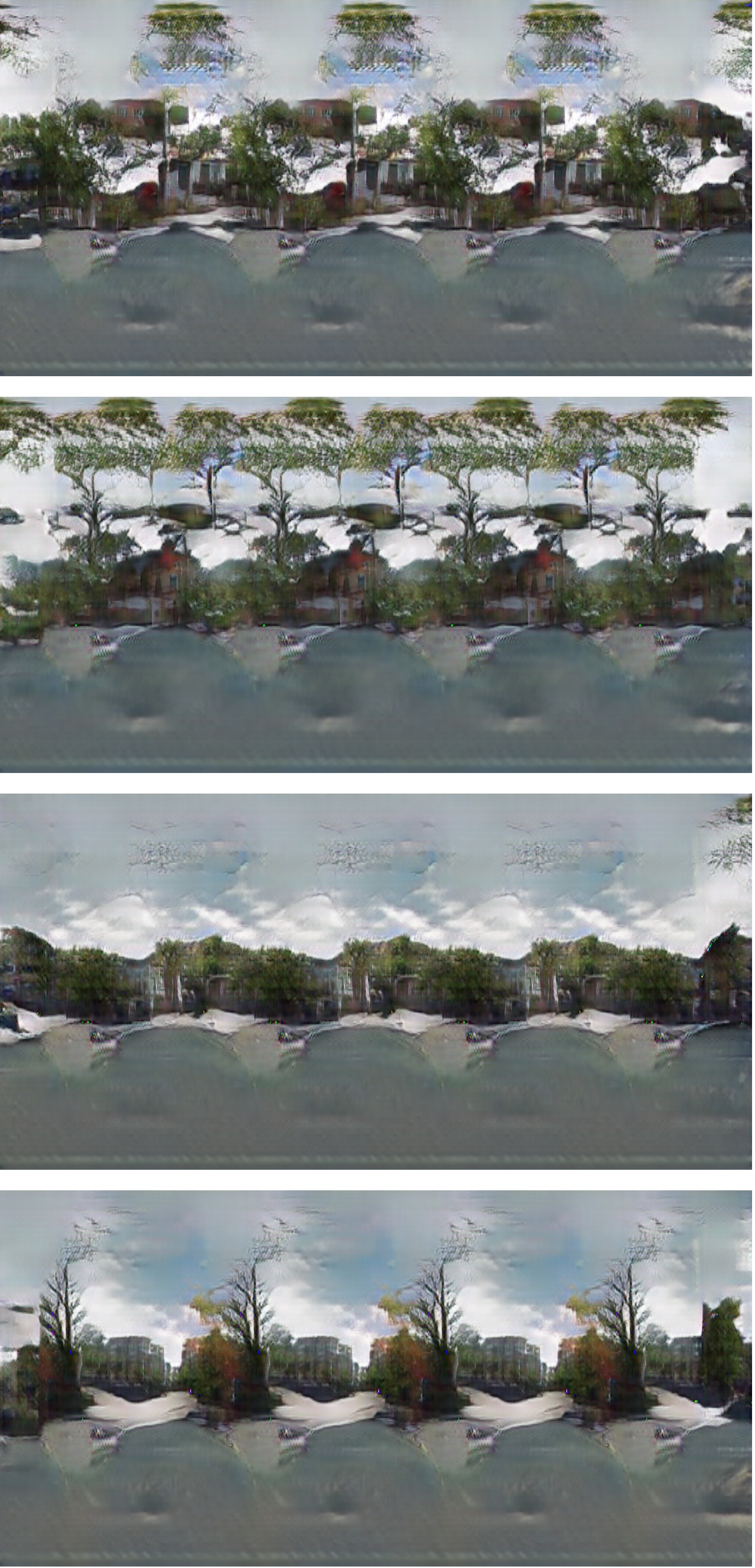}
        \caption{Pix2Pix}
        \label{fig:1b}
    \end{subfigure}
    %
    \begin{subfigure}[b]{0.176\linewidth}
        \includegraphics[width=\linewidth]{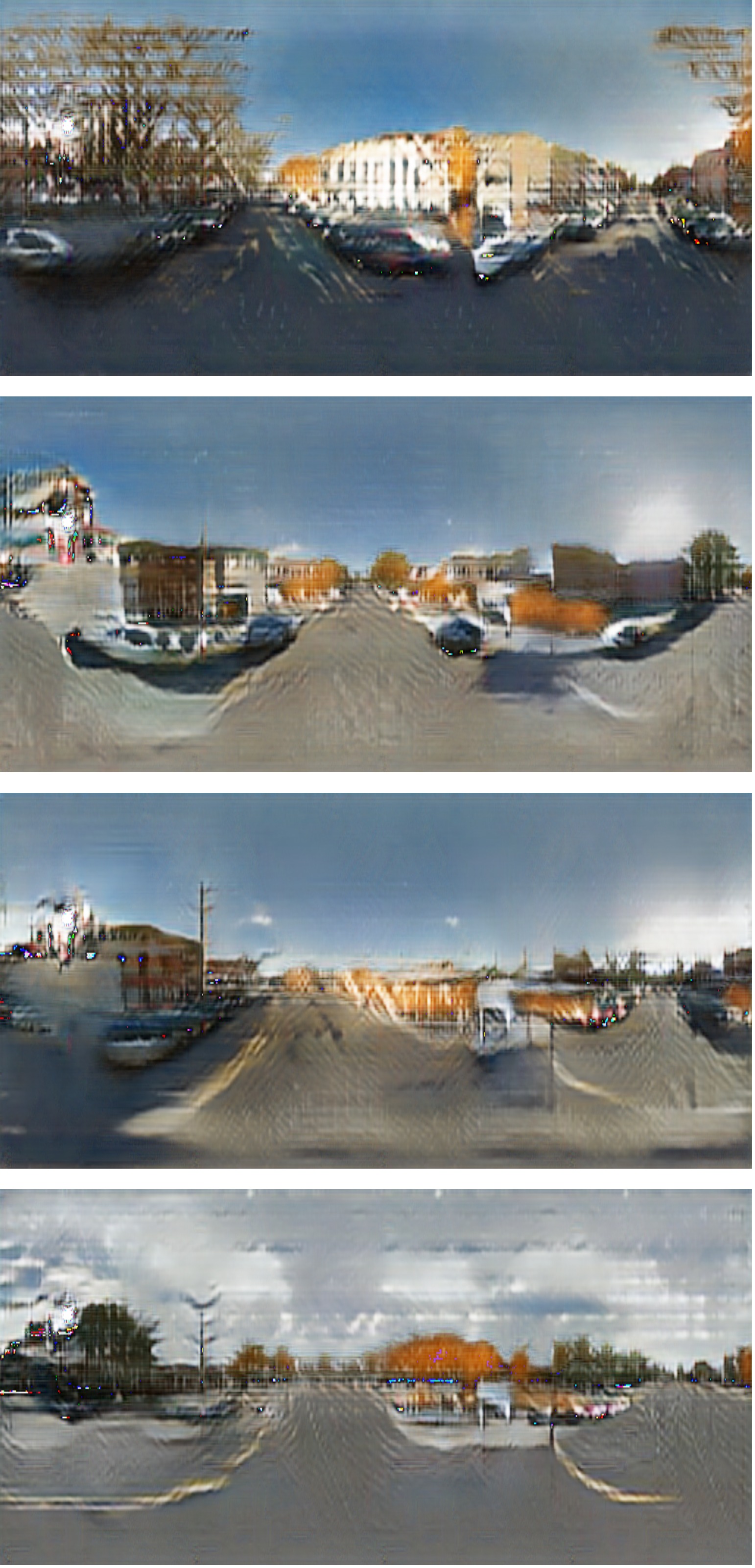}
        \caption{PanoGAN}
        \label{fig:1c}
    \end{subfigure}
    %
    \begin{subfigure}[b]{0.176\linewidth}
        \includegraphics[width=\linewidth]{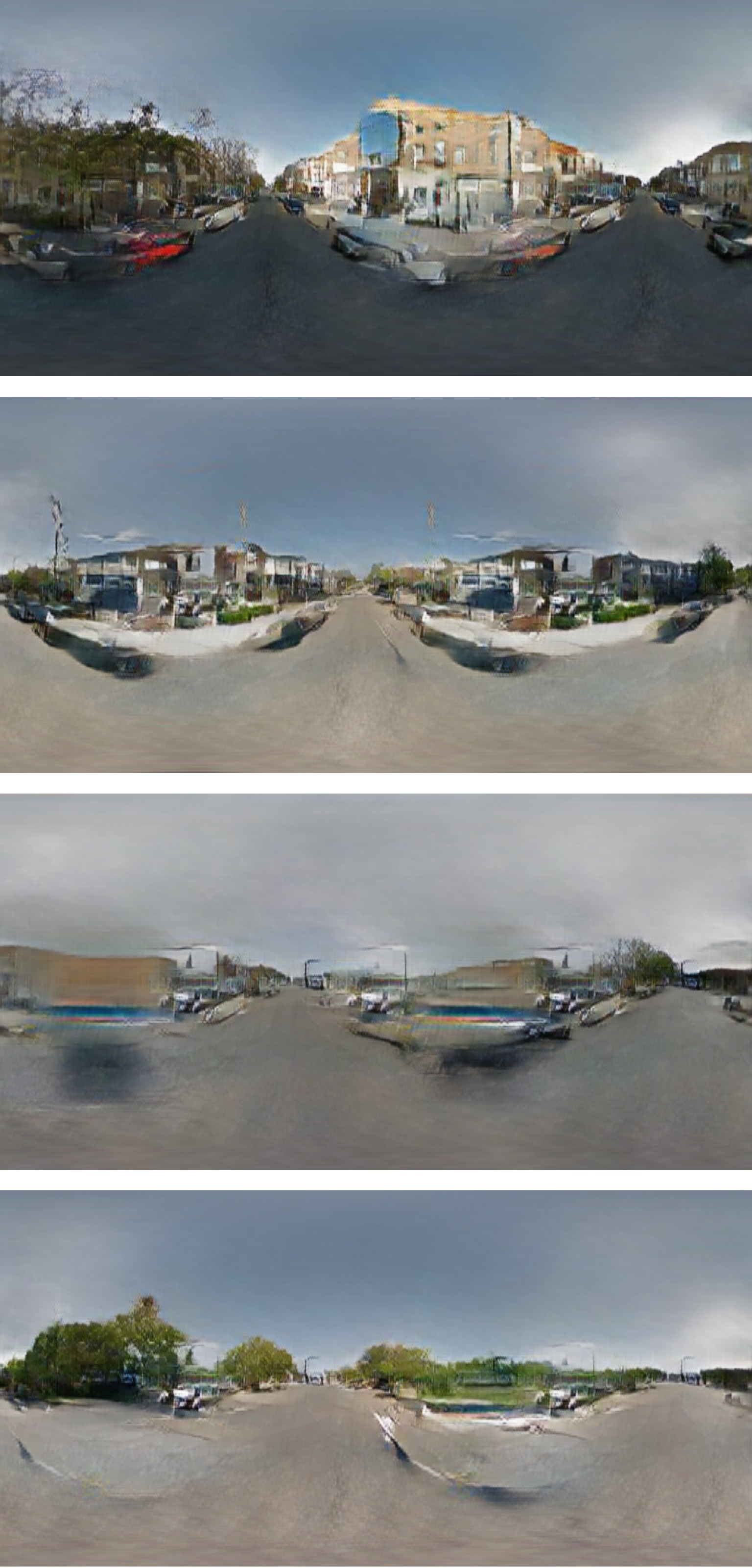}
        \caption{Sat2Density}
        \label{fig:1d}
    \end{subfigure}
    %
    \begin{subfigure}[b]{0.176\linewidth}
        \includegraphics[width=\linewidth]{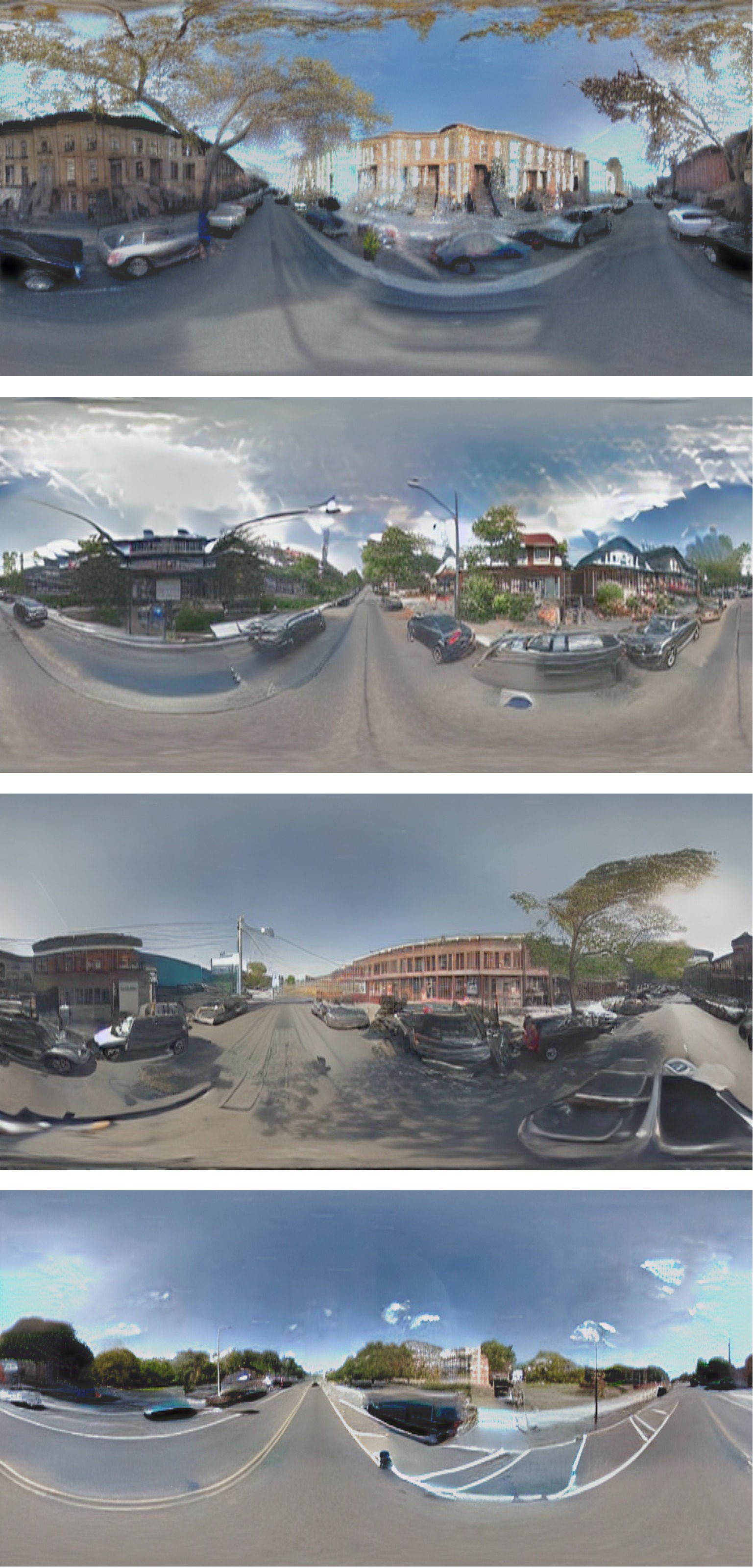}
        \caption{Ours}
        \label{fig:1e}
    \end{subfigure}
    %
    \begin{subfigure}[b]{0.176\linewidth}
        \includegraphics[width=\linewidth]{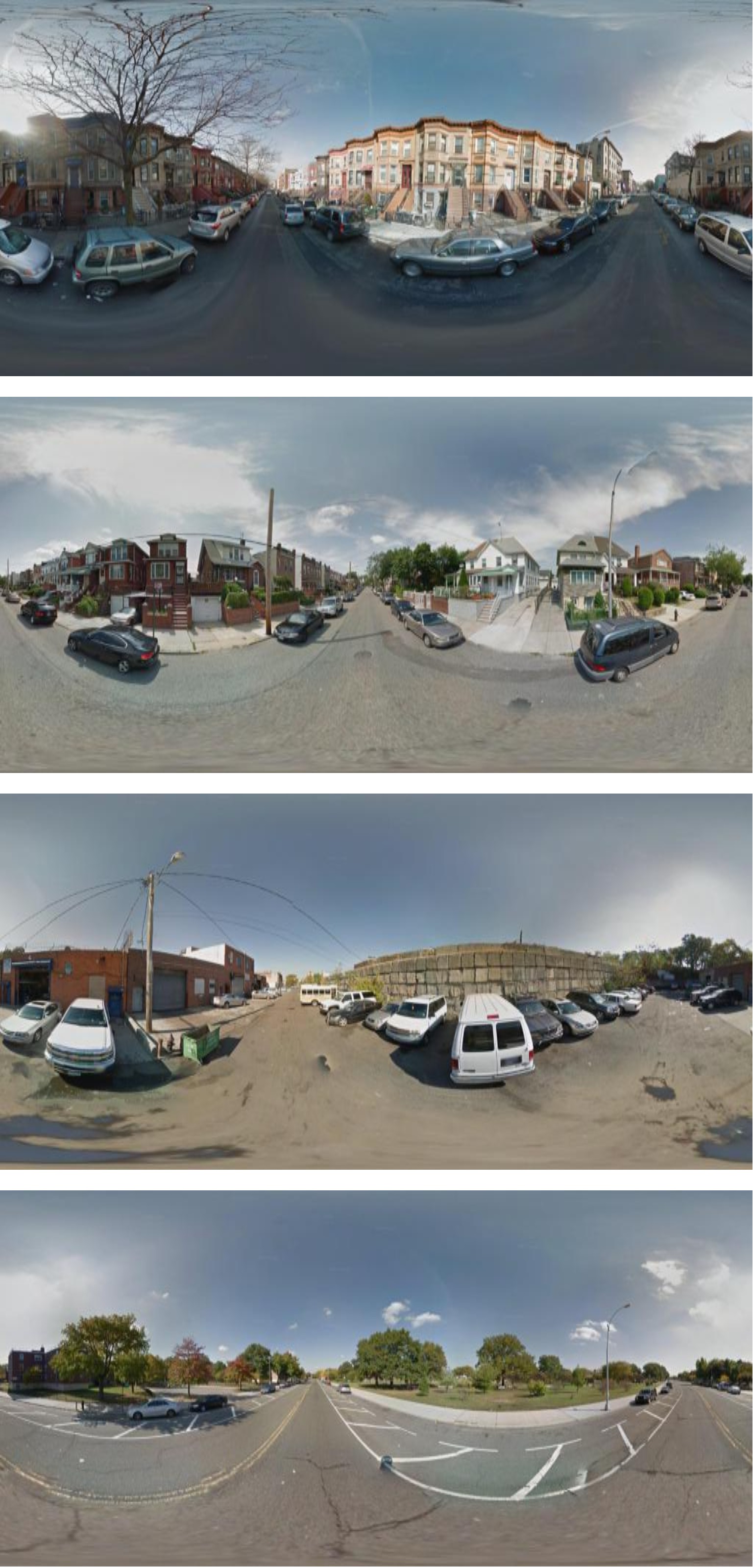}
        \caption{Ground Truth}
        \label{fig:1f}
    \end{subfigure}
    
    \caption{Qualitative results versus baselines. The cross-view synthesis methods that we compare with are trained on our collected center-aligned satellite images. Our approach, which integrates nearby street-level panoramas, not only generates more realistic results when compared to baselines, but more accurate results both semantically and geometrically when compared to the ground truth.}
    \label{fig:results}
\end{figure*}

\begin{figure}[t!]
    \centering
    \includegraphics[width=\linewidth]{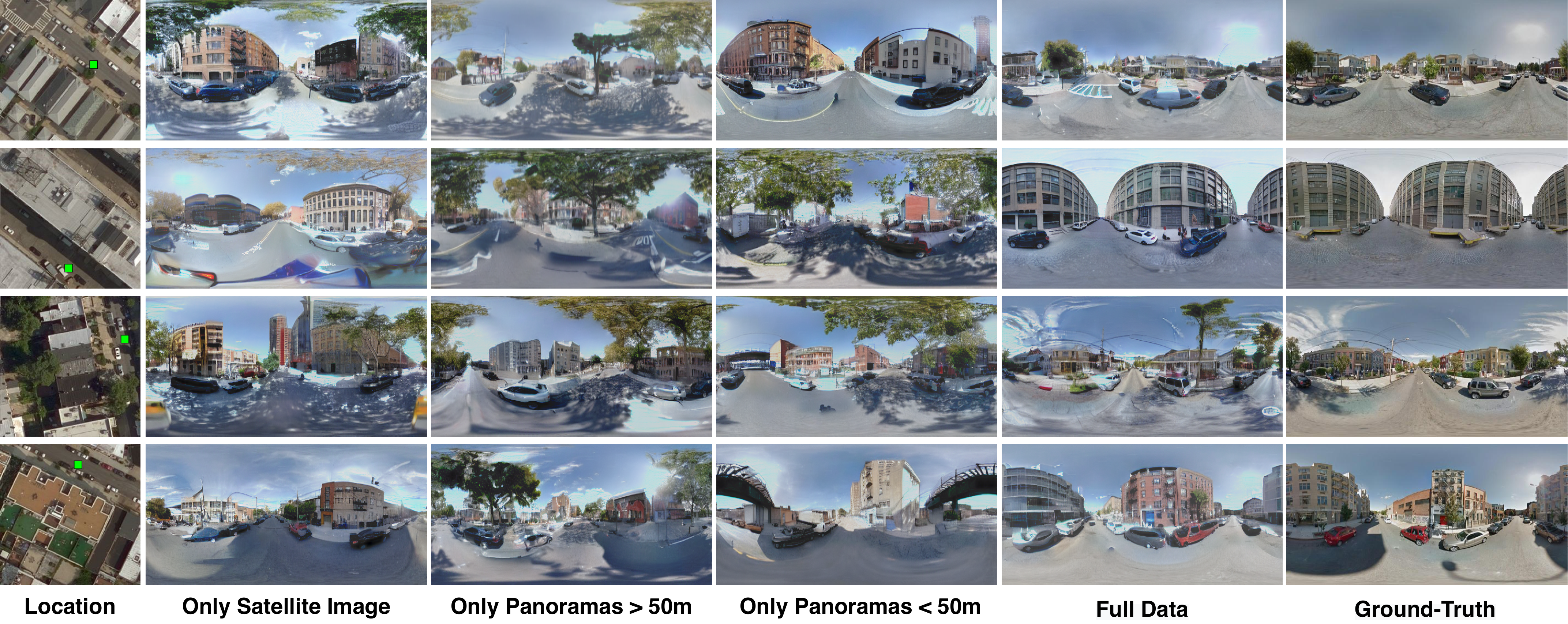}
    \caption{Qualitative results for MVPS using different sources of data.}
    \label{fig:vis_data_ablation}
\end{figure}

\begin{table*}[t!]
\footnotesize
\caption{Comparison with cross-view synthesis methods on Brooklyn test set. Center-Aligned: use satellite data that are center-aligned with the target location.}
    \begin{center}
\begin{tabular}{@{}c|l|cccccc@{}}
\toprule
Center-Aligned                     & Method        & PSNR↑          & SSIM↑           & $P_{alex}$↓    & RMSE↓          & FID↓           & SD↑            \\ \midrule
                & Pix2pix~\cite{isola2017image}       & 11.93          & 0.0950           & 0.6161          & 64.75          & 413.29         & 9.07          \\
                             & PanoGAN~\cite{wu2022cross}       & 13.10          & 0.2981          & 0.5583          & 56.98          & 166.30          & 12.04        \\
   \XSolidBrush             & Sat2density~\cite{qian2023sat2density}   & 13.39          & 0.4325          & 0.5407          & 55.38          & 153.32         & 13.22          \\
                             & GeoDiffusion \textbf{(Ours)} & \textbf{14.14} & \textbf{0.4329} & \textbf{0.4343} & \textbf{51.42} & \textbf{33.68} & \textbf{13.60} \\ \midrule
   & Pix2pix~\cite{isola2017image}       & 12.19          & 0.3375          & 0.5503          & 63.14          & 263.97         & 11.71          \\
                             & PanoGAN~\cite{wu2022cross}       & 13.64          & 0.4044          & 0.4856          & 53.71          & 130.98         & 13.28          \\
    \Checkmark                  & Sat2Density~\cite{qian2023sat2density}   & 14.57 & 0.4465 & 0.4684          & 48.54 & 87.77          & 13.66          \\
                             & GeoDiffusion \textbf{(Ours)} & \textbf{14.66}          & \textbf{0.4498}          & \textbf{0.4206}  & 
                             \textbf{48.31}         & \textbf{31.07} 
                             & \textbf{13.79} \\ \bottomrule
\end{tabular}
\end{center}
    \label{tbl:comparison}
\end{table*}

\begin{table*}[t!]
\footnotesize
\caption{Cross dataset evaluation on Queens. Center-Aligned: use satellite data that are center-aligned with the target location.}
    \begin{center}
\begin{tabular}{@{}c|l|cccccc@{}}
\toprule
Center-Aligned                     & Method        & PSNR↑          & SSIM↑           & $P_{alex}$↓    & RMSE↓          & FID↓           & SD↑            \\ \midrule
                & Pix2pix~\cite{isola2017image}       & 11.82& 0.1719& 0.6435& 65.74& 386.41& 9.51\\
                             & PanoGAN~\cite{wu2022cross}       & 12.79& 0.2968& 0.5770& 59.24& 183.82& 12.02\\
   \XSolidBrush             & Sat2Density~\cite{qian2023sat2density}   & 12.93& 0.4023& 0.5524& 58.63& 181.45& 13.31\\
                             & GeoDiffusion \textbf{(Ours)} & \textbf{13.54}& \textbf{0.4239}& \textbf{0.4661}& \textbf{55.80}& \textbf{57.66}& \textbf{13.55}\\ \midrule
   & Pix2pix~\cite{isola2017image}       & 11.77& 0.2537& 0.5450& 66.73& 265.53& 10.83\\
                             & PanoGAN~\cite{wu2022cross}       & 13.45& 0.4014& 0.4933& 57.01& 99.56& 13.08\\
    \Checkmark                  & Sat2Density~\cite{qian2023sat2density}   & 13.87& 0.4478& 0.4835& 53.76& 107.14& 13.61\\
                             & GeoDiffusion \textbf{(Ours)} & \textbf{14.08}& \textbf{0.4488}& \textbf{0.4585}& \textbf{53.46}& \textbf{54.26}& \textbf{13.72}\\ \bottomrule
\end{tabular}
\end{center}
    \label{tbl:comparison queens}
\end{table*}

\begin{table}[t!]   
    \footnotesize
    \caption{Ablation study for geospatial attention. Local: local-level attention. Global: global-level attention.}
    \begin{center}
        \renewcommand{\arraystretch}{1.0} 
        \begin{tabular}{@{}cc|cccccc@{}}
            \toprule
            Local & Global & PSNR↑ & SSIM↑ & $P_{alex}$↓ & RMSE↓ & FID↓ & SD↑ \\ 
            \midrule
                  &                & 11.69   & 0.3567     & 0.5089            & 67.41     & 52.56   & 12.60   \\
            \Checkmark       &     & 12.95  & 0.3757  & 0.4686  & 58.49 & 37.90  & 12.79    \\
            & \Checkmark    & 12.83  & 0.3898  & 0.4757   & 59.45  & 35.54 & 12.89 \\
            \Checkmark   & \Checkmark  &  \textbf{14.14 }  & \textbf{0.4329}    & \textbf{0.4343}              & \textbf{51.42}  & \textbf{33.68}& \textbf{13.60} \\ \bottomrule
        \end{tabular}
    \end{center}
    \label{tbl:attention_ablation}
\end{table}

\paragraph{\textbf{Implementation Details}}
We use a pretrained Stable Diffusion~\cite{Rombach_2022_CVPR} model (v1.5) with the default parameters. For optimization, we use AdamW with a learning rate of $\lambda=2\times10^{-5}$. The input images are resized to $256\times1024$ as local conditions. Specifically, the satellite image is resized to $256\times256$ and replicated horizontally four times.  We adopt ViT-Adapter~\cite{chen2022vision} to get the segmentation map of the target panorama. We use DDIM~\cite{song2020denoising} for sampling with the number of time steps set to 50. The classifier free guidance~\cite{ho2022classifier} is set to 7.5. During training, the segmentation map of the target image, the satellite image, and a set of nearby street-level panoramas are passed through the controllable diffusion model to synthesize the target panorama. At inference, only the satellite image and available nearby panoramas are required to synthesize the target panorama image, following the same assumption made in~\cite{regmi2018cross} since segmentation maps are unlikely to be available in the real world. 

\paragraph{\textbf{Modality Dropout}} Using multi-conditioned inputs results in the high reliance on one or a subset of the conditions. To avoid this, we implement a modality dropout strategy in three forms: 1) randomly omit each individual condition 2) randomly keep all conditions and 3) randomly drop all conditions. This allows the model to learn the mixed-view panorama synthesis task based on arbitrary conditions, reducing the model's reliance on individual conditions, and helping learn the relationships between different modality compositions. In our experiments, during training, we set the rate to keep/drop all conditions as 0.3 and 0.1 respectively, and set the dropout rate of each condition to 0.1. For the text prompts, we randomly replace 50\% of text prompts with empty strings to enhance the model's ability to learn image geometric relationships.

\subsection{Quantitative Results}

We compare the performance of our method against several baseline methods in \tabref{comparison} and cross-domain evaluation results in \tabref{comparison queens}, using the Brooklyn and Queens dataset. We also show qualitative results in \figref{results}. For the baseline methods, Pix2pix~\cite{isola2017image} is a traditional GAN-based image-to-image translation method. PanoGAN~\cite{wu2022cross} is a recent GAN-based cross-view synthesis method. Sat2Density~\cite{qian2023sat2density} is the state-of-the-art cross-view synthesis method. 

As mentioned in previous works~\cite{Zhu_2021_CVPR}, in practical real-world applications, the panorama that users want to synthesis can occur at arbitrary locations in the area of interest, in this case perfectly aligned correspondence is not guaranteed. We both report the results on (1) the original Brooklyn and Queens dataset, in which satellite image and target location are not center-aligned and (2) our collected satellite images that are center-aligned with the target location. In both these two circumstances, our method outperforms all the cross-view synthesis methods. Specifically, our method shows more superiority on unaligned satellite data, while other cross-view synthesis methods have a significant performance degradation on unaligned satellite data. Our diffusion-based method has advantages especially in high-level metrics LPIPS and FID scores, showing its ability to generate high-fidelity images with global-level accuracy. Overall, our method shows more flexibility as it can achieve competitive performance without requiring the target panorama to be located at the center of the satellite image, improving the efficiency especially in parallelly generating multiple panoramas for a given region and bridging the gap between current research and practical applications.

\begin{table}[t!]
    \footnotesize
    \caption{Ablation study on data distribution. We set the threshold between `near' and `far' to 50m. S: Satellite image P: Panoramas.}
    \begin{center}
        \renewcommand{\arraystretch}{1.0} 
        \begin{tabular}{@{}ccc|cccccc@{}}
            \toprule
            $S$ &  $P \geq 50m$  &  $P \textless 50m$  & PSNR↑          & SSIM↑           & $P_{alex}$↓    & RMSE↓          & FID↓           & SD↑            \\ 
            \midrule
            \Checkmark      &                &               & 11.85          & 0.3143          & 0.5416          & 65.67          & 48.68          & 12.26          \\
                             & \Checkmark        &               & 12.20           & 0.3229          & 0.5244          & 63.10           & 42.81          & 12.37          \\
                             &                & \Checkmark    & 12.44          & 0.3224          & 0.5285          & 63.40           & 42.94          & 12.49          \\
            \Checkmark    & \Checkmark     &               & 12.67          & 0.3488          & 0.5174          & 57.12          & 39.71         & 12.61          \\
            \Checkmark    &      &  \Checkmark    & 13.42          & 0.4136          & 0.5013          & 55.01          & 38.38         & 13.51          \\
            \Checkmark      & \Checkmark    & \Checkmark    & \textbf{14.14} & \textbf{0.4329} & \textbf{0.4343} & \textbf{51.42} & \textbf{33.68} & \textbf{13.60}\\ 
            \bottomrule
        \end{tabular}
    \end{center}
    \label{tbl:data_ablation}
\end{table}


\begin{table}[t!]%
\centering
\footnotesize
\caption{Ablation study. We analyze the effect of modality dropout and adding segmentation maps during the training stage.}
\setlength{\tabcolsep}{1.0mm}\begin{tabular}{@{}c|cccccc@{}}
\toprule
Ablation Objective         & PSNR↑ & SSIM↑ & $P_{alex}$↓ & RMSE↓ & FID↓ & SD↑ \\ \midrule
w/o modality dropout &      11.02&      0.3215&        0.5764&      72.31&     185.27&    12.84\\
w/o segmentation maps in training &  12.54 &  0.3665 &   0.5267 &   64.46 &     96.87&    12.98 \\
Simple attention w/o geospatial attention adapter &	11.83 &	0.3275	& 0.5783	&72.85	& 98.86	& 12.75 \\
Concat conditions and noise along spatial dimensions	& 13.12	& 0.3589	& 0.5172	& 58.23	& 82.36	& 13.16  \\
Simple attention w/o geospatial information &  13.31 &  0.3896 &   0.4765 &   55.27 &     56.83 &    13.35 \\
Full             & \textbf{14.14} & \textbf{0.4329} & \textbf{0.4343} & \textbf{51.42} & \textbf{33.68} & \textbf{13.60} \\ \bottomrule
\end{tabular}
\label{tbl:dropout}
\end{table}

\subsection{Ablation Study}

We conduct an ablation study that evaluates the impact of individual components of our proposed geospatial attention adapter on the resulting image quality. The results are shown in \tabref{attention_ablation}. Experimental results show that both the local and global geospatial attention can guide the diffusion model to better utilize the geometric relationships between mixed-view modalities. Furthermore, combining local and global geospatial attention helps supervise the model towards generating results with accurate layout distributions.  

We also analyze the effect of the satellite images and near panoramas of different distances on our model. The results are shown in \tabref{data_ablation} and \figref{vis_data_ablation}. In this experiment, we set thresholds to 50 meters based on the haversine distance between the near panorama and the target panorama to define `near' and `far'. We train our model on each subset and all available data.  Panoramas that are near the target location have larger effect on the synthesis quality as there are more overlapping regions. However, the model does not rely on the `near' panoramas to capture accurate geometry, since the model trained on `far' panoramas and satellite images outperforms the model only trained on `near' panoramas. The model trained on a subset of the data cannot synthesize accurate layout and sometimes confuses the sky region with trees. In summary, using all data significantly improves the synthesis quality of the model. 

We conduct further ablation study in \tabref{dropout}. We first conduct an experiment on modality dropout strategy. Compared with keeping all the input images, adding modality dropout achieves \textbf{24.6\%↓} in $P_{alex}$ and \textbf{81.8\%↓} in FID. We also remove the segmentation map branch of the multi-conditioned diffusion model in the training stage, which leads to severe performance degregation. Furthermore, we conduct experiments about: (1)concatenate conditions along the channel dimension and train a ControlNet to guide generation; (2) concatenate conditions and noise along spatial dimensions, followed by fine-tuning the diffusion model and (3) use simple attention to compute the attention map between the source view image and target view image, without adding location information and orientation information. All of these models show worse performance. These experiments demonstrate the effectiveness of modality dropout strategy and the supervision of segmentation maps in the training stage.


\begin{figure}[t!]
    \centering
    \includegraphics[width=\linewidth]{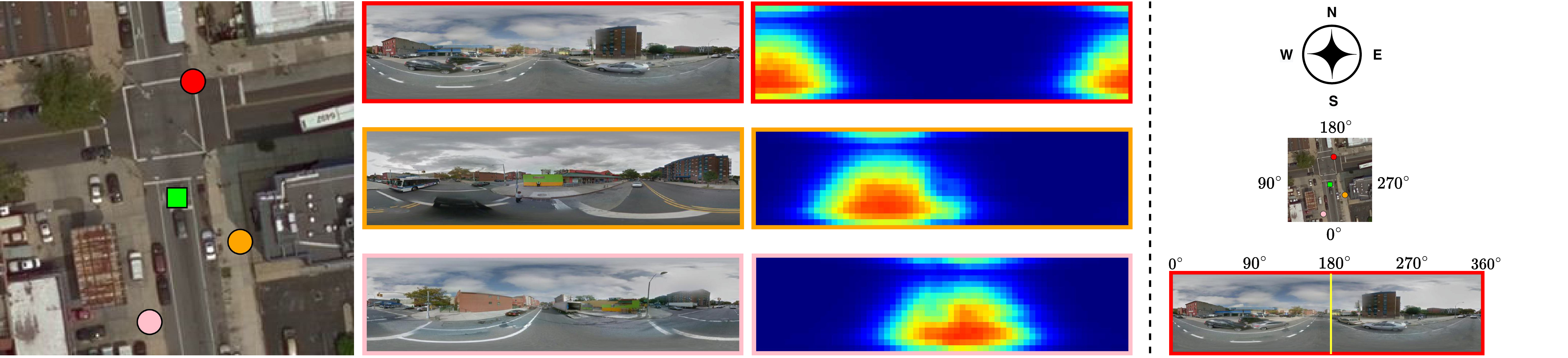}
    \caption{Visualization of local geospatial attention. The target location is represented by a green square in the satellite image. The nearby street-level panoramas (color-coded borders) are represented by same-colored circles in the satellite image.}
    \label{fig:local_att}
\end{figure}

\begin{figure}[t!]
    \centering
    \includegraphics[width=\linewidth]{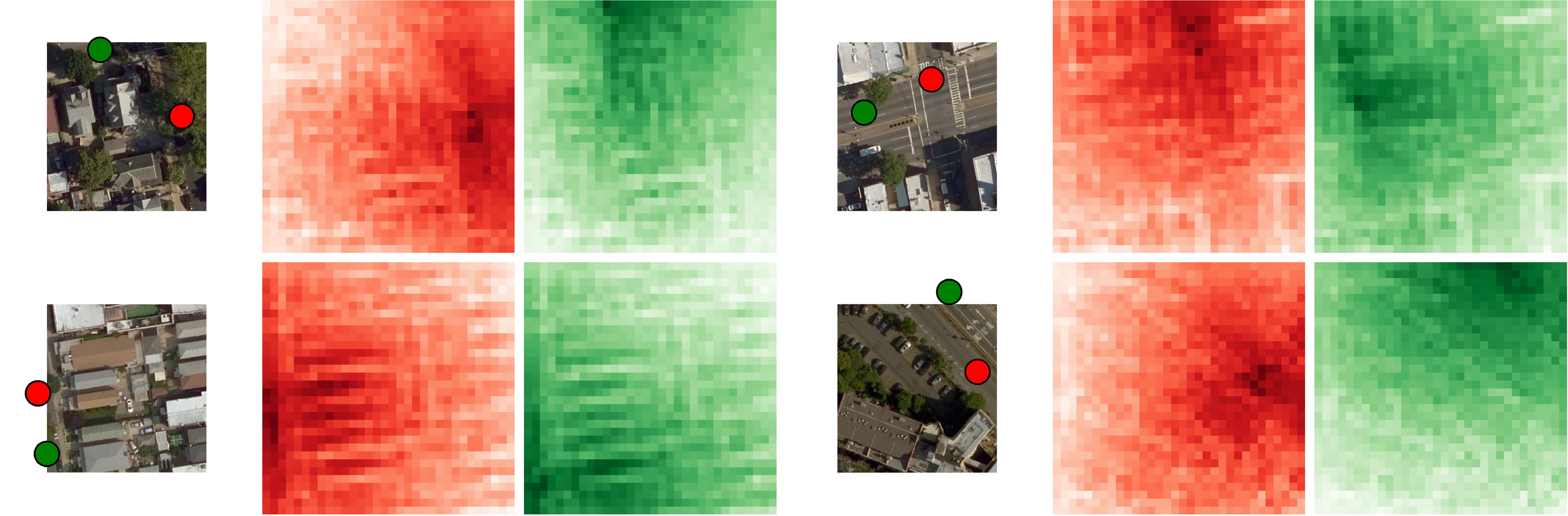}
    \caption{Visualization of global geospatial attention. The color-coded attention maps for two target locations are shown, corresponding to the same-colored dots in the satellite image. Darker colors represent more salient regions.}
    \label{fig:global_att}
\end{figure}

\subsection{Visualization of Geospatial Attention}

We demonstrate the concept of geospatial attention visually. \figref{local_att} shows an example of local geospatial attention, which uses the relative orientation and distance of a nearby street-level panorama in addition to semantic content. The street-level panoramas and paired local-level attention map use an east-north-up coordinate system, i.e., the center points to the north and the left/right boundary points to the south as~\cite{Zhu_2021_CVPR}.  The target location is indicated by the green square in the satellite image and the location of the panoramas is indicated by the same-colored dot. Larger attention values (red) reflect the region in the panoramas that orient towards the target. Note that when the image and target locations are further apart, the high attention region shrinks, essentially reflecting the narrower field of view.

\figref{global_att} shows an example of global geospatial attention for two target synthesis locations. Larger attention values (darker) reflect the regions in the overhead view that contribute the most. Note that, for the Brooklyn and Queens dataset, the street-level panoramas are primarily collected along the streets. This is captured in the global-level attention maps, which tend to show higher attention scores in the street-level regions near the target location (omitting buildings). Additional visualizations are shown in the supplementary material.

\subsection{Discussion about Seamless}

We adopt the Left-Right Consistency Error (LRCE) metric~\citep{shen2022panoformer} to quantitatively assess the consistency along the left-right boundaries by analyzing the horizontal gradients. For depth evaluation, we integrate Depth Anywhere~\citep{wang2024depthanywhere}, a state-of-the-art panorama depth estimation method, to predict the depth maps of the panoramas. We compute the horizontal gradients of both the ground-truth depth maps and those of the generated panoramas, and then calculate the gradient differences. By randomly sampling 100 generated panoramas from the test set, we determined the mean gradient difference, as detailed in \tabref{seamless}. Our experimental results indicate that, compared to previous cross-view synthesis methods, our approach achieves superior left-right boundary consistency, thereby demonstrating its seamless performance.


\begin{table}[]
\centering
\footnotesize
\caption{Quantatitive comparison about Seamless.}
\begin{tabular}{@{}lccc@{}}
\toprule
     & PanoGAN~\cite{wu2022cross} & Sat2Density~\cite{qian2023sat2density} & \textbf{Ours} \\ \midrule
LRCE & 0.0897  &  0.0653    & \textbf{0.0436}     \\ \bottomrule
\end{tabular}
\label{tbl:seamless}
\end{table}

\section{Conclusion and Future Work}

We introduced the task of mixed-view synthesis, which extends the cross-view synthesis task to also include a set of nearby street-level panoramas as input. We proposed a novel multi-conditioned, end-to-end geospatial attention-guided diffusion framework for combining information from all input imagery to guide the diffusion-based synthesis process, achieving geometry-guided fine-grained spatial control. Unlike some cross-view approaches, our approach does not require a segmentation map of the target panorama during the inference stage or, like all previous cross-view approaches, that the target panorama be located at exactly the center of the satellite image. Together, these characteristics dramatically increase flexibility and ease of use. Experimental results demonstrate the effectiveness of our proposed model, in particular its ability to handle situations when the panoramas far from the target location.

For urban scenes, transient objects (e.g., cars, pedestrians) bring challenges to our method. Also due to the inherent randomness of the diffusion models, maintaining the view consistency between adjacent locations is still challenging in some detailed regions. Understanding how to reduce the influence of transient objects on guided diffusion models to synthesize clean and view-consistent panorama sequences would be a future research direction.


\bibliography{main}

\begin{thebibliography}{72}
\providecommand{\natexlab}[1]{#1}
\providecommand{\url}[1]{\texttt{#1}}
\expandafter\ifx\csname urlstyle\endcsname\relax
  \providecommand{\doi}[1]{doi: #1}\else
  \providecommand{\doi}{doi: \begingroup \urlstyle{rm}\Url}\fi

\bibitem[Carion et~al.(2020)Carion, Massa, Synnaeve, Usunier, Kirillov, and Zagoruyko]{carion2020end}
Nicolas Carion, Francisco Massa, Gabriel Synnaeve, Nicolas Usunier, Alexander Kirillov, and Sergey Zagoruyko.
\newblock End-to-end object detection with transformers.
\newblock In \emph{European conference on computer vision}, pp.\  213--229. Springer, 2020.

\bibitem[Chang et~al.(2022)Chang, Zhang, Jiang, Liu, and Freeman]{Chang_2022_CVPR}
Huiwen Chang, Han Zhang, Lu~Jiang, Ce~Liu, and William~T. Freeman.
\newblock Maskgit: Masked generative image transformer.
\newblock In \emph{Proceedings of the IEEE/CVF conference on computer vision and pattern recognition}, 2022.

\bibitem[Chen et~al.(2021)Chen, Fan, and Panda]{Chen_2021_ICCV}
Chun-Fu~(Richard) Chen, Quanfu Fan, and Rameswar Panda.
\newblock Crossvit: Cross-attention multi-scale vision transformer for image classification.
\newblock In \emph{Proceedings of the IEEE/CVF International Conference on Computer Vision (ICCV)}, 2021.

\bibitem[Chen et~al.(2022)Chen, Duan, Wang, He, Lu, Dai, and Qiao]{chen2022vision}
Zhe Chen, Yuchen Duan, Wenhai Wang, Junjun He, Tong Lu, Jifeng Dai, and Yu~Qiao.
\newblock Vision transformer adapter for dense predictions.
\newblock \emph{arXiv preprint arXiv:2205.08534}, 2022.

\bibitem[Dhariwal \& Nichol(2021)Dhariwal and Nichol]{dhariwal2021diffusion}
Prafulla Dhariwal and Alexander Nichol.
\newblock Diffusion models beat gans on image synthesis.
\newblock \emph{Advances in neural information processing systems}, 34:\penalty0 8780--8794, 2021.

\bibitem[Dosovitskiy et~al.(2020)Dosovitskiy, Beyer, Kolesnikov, Weissenborn, Zhai, Unterthiner, Dehghani, Minderer, Heigold, Gelly, et~al.]{dosovitskiy2020image}
Alexey Dosovitskiy, Lucas Beyer, Alexander Kolesnikov, Dirk Weissenborn, Xiaohua Zhai, Thomas Unterthiner, Mostafa Dehghani, Matthias Minderer, Georg Heigold, Sylvain Gelly, et~al.
\newblock An image is worth 16x16 words: Transformers for image recognition at scale.
\newblock \emph{arXiv preprint arXiv:2010.11929}, 2020.

\bibitem[Esser et~al.(2021)Esser, Rombach, and Ommer]{esser2021taming}
Patrick Esser, Robin Rombach, and Bjorn Ommer.
\newblock Taming transformers for high-resolution image synthesis.
\newblock In \emph{Proceedings of the IEEE/CVF conference on computer vision and pattern recognition}, 2021.

\bibitem[Fukui et~al.(2019)Fukui, Hirakawa, Yamashita, and Fujiyoshi]{Fukui_2019_CVPR}
Hiroshi Fukui, Tsubasa Hirakawa, Takayoshi Yamashita, and Hironobu Fujiyoshi.
\newblock Attention branch network: Learning of attention mechanism for visual explanation.
\newblock In \emph{Proceedings of the IEEE/CVF conference on computer vision and pattern recognition}, 2019.

\bibitem[He et~al.(2020)He, Yan, Fragkiadaki, and Yu]{he2020epipolar}
Yihui He, Rui Yan, Katerina Fragkiadaki, and Shoou-I Yu.
\newblock Epipolar transformers.
\newblock In \emph{Proceedings of the IEEE/CVF conference on computer vision and pattern recognition}, 2020.

\bibitem[Heusel et~al.(2017)Heusel, Ramsauer, Unterthiner, Nessler, and Hochreiter]{heusel2017gans}
Martin Heusel, Hubert Ramsauer, Thomas Unterthiner, Bernhard Nessler, and Sepp Hochreiter.
\newblock Gans trained by a two time-scale update rule converge to a local nash equilibrium.
\newblock In \emph{Advances in neural information processing systems}, 2017.

\bibitem[Ho \& Salimans(2022)Ho and Salimans]{ho2022classifier}
Jonathan Ho and Tim Salimans.
\newblock Classifier-free diffusion guidance.
\newblock \emph{arXiv preprint arXiv:2207.12598}, 2022.

\bibitem[Hu et~al.(2018{\natexlab{a}})Hu, Shen, Albanie, Sun, and Vedaldi]{hu2018gather}
Jie Hu, Li~Shen, Samuel Albanie, Gang Sun, and Andrea Vedaldi.
\newblock Gather-excite: Exploiting feature context in convolutional neural networks.
\newblock In \emph{Advances in neural information processing systems}, 2018{\natexlab{a}}.

\bibitem[Hu et~al.(2018{\natexlab{b}})Hu, Shen, and Sun]{hu2018squeeze}
Jie Hu, Li~Shen, and Gang Sun.
\newblock Squeeze-and-excitation networks.
\newblock In \emph{Proceedings of the IEEE/CVF conference on computer vision and pattern recognition}, 2018{\natexlab{b}}.

\bibitem[Isola et~al.(2017)Isola, Zhu, Zhou, and Efros]{isola2017image}
Phillip Isola, Jun-Yan Zhu, Tinghui Zhou, and Alexei~A Efros.
\newblock Image-to-image translation with conditional adversarial networks.
\newblock In \emph{Proceedings of the IEEE/CVF conference on computer vision and pattern recognition}, 2017.

\bibitem[Jaderberg et~al.(2015)Jaderberg, Simonyan, Zisserman, et~al.]{jaderberg2015spatial}
Max Jaderberg, Karen Simonyan, Andrew Zisserman, et~al.
\newblock Spatial transformer networks.
\newblock In \emph{Advances in neural information processing systems}, 2015.

\bibitem[Kerbl et~al.(2023)Kerbl, Kopanas, Leimk{\"u}hler, and Drettakis]{kerbl3Dgaussians}
Bernhard Kerbl, Georgios Kopanas, Thomas Leimk{\"u}hler, and George Drettakis.
\newblock 3d gaussian splatting for real-time radiance field rendering.
\newblock \emph{ACM Transactions on Graphics}, 42\penalty0 (4), July 2023.

\bibitem[Krizhevsky et~al.(2012)Krizhevsky, Sutskever, and Hinton]{krizhevsky2012imagenet}
Alex Krizhevsky, Ilya Sutskever, and Geoffrey~E Hinton.
\newblock Imagenet classification with deep convolutional neural networks.
\newblock In \emph{Advances in neural information processing systems}, 2012.

\bibitem[Li et~al.(2023)Li, Liu, Wu, Mu, Yang, Gao, Li, and Lee]{li2023gligen}
Yuheng Li, Haotian Liu, Qingyang Wu, Fangzhou Mu, Jianwei Yang, Jianfeng Gao, Chunyuan Li, and Yong~Jae Lee.
\newblock Gligen: Open-set grounded text-to-image generation.
\newblock In \emph{Proceedings of the IEEE/CVF conference on computer vision and pattern recognition}, pp.\  22511--22521, 2023.

\bibitem[Li et~al.(2021)Li, Li, Cui, Qin, Pollefeys, and Oswald]{li2021sat2vid}
Zuoyue Li, Zhenqiang Li, Zhaopeng Cui, Rongjun Qin, Marc Pollefeys, and Martin~R Oswald.
\newblock Sat2vid: street-view panoramic video synthesis from a single satellite image.
\newblock In \emph{Proceedings of the IEEE/CVF International Conference on Computer Vision (ICCV)}, 2021.

\bibitem[Li et~al.(2024)Li, Li, Cui, Pollefeys, and Oswald]{li2024sat2scene}
Zuoyue Li, Zhenqiang Li, Zhaopeng Cui, Marc Pollefeys, and Martin~R. Oswald.
\newblock Sat2scene: 3d urban scene generation from satellite images with diffusion.
\newblock In \emph{Proceedings of the IEEE/CVF Conference on Computer Vision and Pattern Recognition (CVPR)}, pp.\  7141--7150, June 2024.

\bibitem[Liao et~al.(2024)Liao, Xu, Lin, Ren, Wei, and Zhao]{cylin_painting}
Kang Liao, Xiangyu Xu, Chunyu Lin, Wenqi Ren, Yunchao Wei, and Yao Zhao.
\newblock Cylin-painting: Seamless 360° panoramic image outpainting and beyond.
\newblock \emph{IEEE Transactions on Image Processing}, 33:\penalty0 382--394, 2024.

\bibitem[Liu \& Li(2019)Liu and Li]{liu2019lending}
Liu Liu and Hongdong Li.
\newblock Lending orientation to neural networks for cross-view geo-localization.
\newblock In \emph{Proceedings of the IEEE/CVF conference on computer vision and pattern recognition}, 2019.

\bibitem[Liu et~al.(2017)Liu, Breuel, and Kautz]{liu2017unsupervised}
Ming-Yu Liu, Thomas Breuel, and Jan Kautz.
\newblock Unsupervised image-to-image translation networks.
\newblock In \emph{Advances in neural information processing systems}, 2017.

\bibitem[Liu et~al.(2024)Liu, Luo, Fan, Wang, Peng, and Zhang]{liu2024citygaussian}
Yang Liu, Chuanchen Luo, Lue Fan, Naiyan Wang, Junran Peng, and Zhaoxiang Zhang.
\newblock Citygaussian: Real-time high-quality large-scale scene rendering with gaussians.
\newblock In \emph{European Conference on Computer Vision}, pp.\  265--282. Springer, 2024.

\bibitem[Martin-Brualla et~al.(2021)Martin-Brualla, Radwan, Sajjadi, Barron, Dosovitskiy, and Duckworth]{martin2021nerf}
Ricardo Martin-Brualla, Noha Radwan, Mehdi~SM Sajjadi, Jonathan~T Barron, Alexey Dosovitskiy, and Daniel Duckworth.
\newblock Nerf in the wild: Neural radiance fields for unconstrained photo collections.
\newblock In \emph{Proceedings of the IEEE/CVF conference on computer vision and pattern recognition}, pp.\  7210--7219, 2021.

\bibitem[Meng et~al.(2019)Meng, Zhao, Chang, Huang, Sun, Tung, and Sigal]{meng2019interpretable}
Lili Meng, Bo~Zhao, Bo~Chang, Gao Huang, Wei Sun, Frederick Tung, and Leonid Sigal.
\newblock Interpretable spatio-temporal attention for video action recognition.
\newblock In \emph{IEEE/CVF International Conference on Computer Vision Workshops}, 2019.

\bibitem[Mildenhall et~al.(2021)Mildenhall, Srinivasan, Tancik, Barron, Ramamoorthi, and Ng]{mildenhall2021nerf}
Ben Mildenhall, Pratul~P Srinivasan, Matthew Tancik, Jonathan~T Barron, Ravi Ramamoorthi, and Ren Ng.
\newblock Nerf: Representing scenes as neural radiance fields for view synthesis.
\newblock \emph{Communications of the ACM}, 65\penalty0 (1):\penalty0 99--106, 2021.

\bibitem[Mnih et~al.(2014)Mnih, Heess, Graves, et~al.]{mnih2014recurrent}
Volodymyr Mnih, Nicolas Heess, Alex Graves, et~al.
\newblock Recurrent models of visual attention.
\newblock In \emph{Advances in neural information processing systems}, 2014.

\bibitem[Mou et~al.(2024)Mou, Wang, Xie, Wu, Zhang, Qi, and Shan]{mou2023t2i}
Chong Mou, Xintao Wang, Liangbin Xie, Yanze Wu, Jian Zhang, Zhongang Qi, and Ying Shan.
\newblock T2i-adapter: Learning adapters to dig out more controllable ability for text-to-image diffusion models.
\newblock In \emph{Proceedings of the AAAI conference on artificial intelligence}, volume~38, pp.\  4296--4304, 2024.

\bibitem[Nichol et~al.(2021)Nichol, Dhariwal, Ramesh, Shyam, Mishkin, McGrew, Sutskever, and Chen]{nichol2021glide}
Alex Nichol, Prafulla Dhariwal, Aditya Ramesh, Pranav Shyam, Pamela Mishkin, Bob McGrew, Ilya Sutskever, and Mark Chen.
\newblock Glide: Towards photorealistic image generation and editing with text-guided diffusion models.
\newblock \emph{arXiv preprint arXiv:2112.10741}, 2021.

\bibitem[Park et~al.(2019)Park, Liu, Wang, and Zhu]{park2019semantic}
Taesung Park, Ming-Yu Liu, Ting-Chun Wang, and Jun-Yan Zhu.
\newblock Semantic image synthesis with spatially-adaptive normalization.
\newblock In \emph{Proceedings of the IEEE/CVF conference on computer vision and pattern recognition}, pp.\  2337--2346, 2019.

\bibitem[Parmar et~al.(2023)Parmar, Kumar~Singh, Zhang, Li, Lu, and Zhu]{parmar2023zero}
Gaurav Parmar, Krishna Kumar~Singh, Richard Zhang, Yijun Li, Jingwan Lu, and Jun-Yan Zhu.
\newblock Zero-shot image-to-image translation.
\newblock In \emph{ACM SIGGRAPH}, 2023.

\bibitem[Qian et~al.(2023)Qian, Xiong, Xia, and Xue]{qian2023sat2density}
Ming Qian, Jincheng Xiong, Gui-Song Xia, and Nan Xue.
\newblock Sat2density: Faithful density learning from satellite-ground image pairs.
\newblock In \emph{Proceedings of the IEEE/CVF International Conference on Computer Vision}, pp.\  3683--3692, 2023.

\bibitem[Qin et~al.(2023)Qin, Zhang, Yu, Feng, Yang, Zhou, Wang, Niebles, Xiong, Savarese, et~al.]{qin2023unicontrol}
Can Qin, Shu Zhang, Ning Yu, Yihao Feng, Xinyi Yang, Yingbo Zhou, Huan Wang, Juan~Carlos Niebles, Caiming Xiong, Silvio Savarese, et~al.
\newblock Unicontrol: A unified diffusion model for controllable visual generation in the wild.
\newblock \emph{arXiv preprint arXiv:2305.11147}, 2023.

\bibitem[Qu et~al.(2023)Qu, Wu, Fei, Nie, and Chua]{qu2023layoutllm}
Leigang Qu, Shengqiong Wu, Hao Fei, Liqiang Nie, and Tat-Seng Chua.
\newblock Layoutllm-t2i: Eliciting layout guidance from llm for text-to-image generation.
\newblock In \emph{Proceedings of the 31st ACM International Conference on Multimedia}, pp.\  643--654, 2023.

\bibitem[Ramesh et~al.(2022)Ramesh, Dhariwal, Nichol, Chu, and Chen]{ramesh2022hierarchical}
Aditya Ramesh, Prafulla Dhariwal, Alex Nichol, Casey Chu, and Mark Chen.
\newblock Hierarchical text-conditional image generation with clip latents.
\newblock \emph{arXiv preprint arXiv:2204.06125}, 1\penalty0 (2):\penalty0 3, 2022.

\bibitem[Regmi \& Borji(2018)Regmi and Borji]{regmi2018cross}
Krishna Regmi and Ali Borji.
\newblock Cross-view image synthesis using conditional gans.
\newblock In \emph{Proceedings of the IEEE/CVF conference on computer vision and pattern recognition}, 2018.

\bibitem[Regmi \& Borji(2019)Regmi and Borji]{regmi2019cross}
Krishna Regmi and Ali Borji.
\newblock Cross-view image synthesis using geometry-guided conditional gans.
\newblock \emph{Computer Vision and Image Understanding}, 187:\penalty0 102788, 2019.

\bibitem[Regmi \& Shah(2019)Regmi and Shah]{regmi2019bridging}
Krishna Regmi and Mubarak Shah.
\newblock Bridging the domain gap for ground-to-aerial image matching.
\newblock In \emph{Proceedings of the IEEE/CVF International Conference on Computer Vision (ICCV)}, 2019.

\bibitem[Rombach et~al.(2022)Rombach, Blattmann, Lorenz, Esser, and Ommer]{Rombach_2022_CVPR}
Robin Rombach, Andreas Blattmann, Dominik Lorenz, Patrick Esser, and Bj\"orn Ommer.
\newblock High-resolution image synthesis with latent diffusion models.
\newblock In \emph{Proceedings of the IEEE/CVF conference on computer vision and pattern recognition}, 2022.

\bibitem[Shen et~al.(2022)Shen, Lin, Liao, Nie, Zheng, and Zhao]{shen2022panoformer}
Zhijie Shen, Chunyu Lin, Kang Liao, Lang Nie, Zishuo Zheng, and Yao Zhao.
\newblock Panoformer: panorama transformer for indoor 360∘ depth estimation.
\newblock In \emph{European Conference on Computer Vision}, pp.\  195--211. Springer, 2022.

\bibitem[Shi et~al.(2022{\natexlab{a}})Shi, Campbell, Yu, and Li]{9674229}
Yujiao Shi, Dylan Campbell, Xin Yu, and Hongdong Li.
\newblock Geometry-guided street-view panorama synthesis from satellite imagery.
\newblock \emph{IEEE transactions on pattern analysis and machine intelligence}, 44\penalty0 (12):\penalty0 10009--10022, 2022{\natexlab{a}}.
\newblock \doi{10.1109/TPAMI.2022.3140750}.

\bibitem[Shi et~al.(2022{\natexlab{b}})Shi, Yu, Liu, Campbell, Koniusz, and Li]{shi2022accurate}
Yujiao Shi, Xin Yu, Liu Liu, Dylan Campbell, Piotr Koniusz, and Hongdong Li.
\newblock Accurate 3-dof camera geo-localization via ground-to-satellite image matching.
\newblock \emph{IEEE transactions on pattern analysis and machine intelligence}, 45\penalty0 (3):\penalty0 2682--2697, 2022{\natexlab{b}}.

\bibitem[Song et~al.(2020)Song, Meng, and Ermon]{song2020denoising}
Jiaming Song, Chenlin Meng, and Stefano Ermon.
\newblock Denoising diffusion implicit models.
\newblock \emph{arXiv preprint arXiv:2010.02502}, 2020.

\bibitem[Song et~al.(2017)Song, Lan, Xing, Zeng, and Liu]{song2017end}
Sijie Song, Cuiling Lan, Junliang Xing, Wenjun Zeng, and Jiaying Liu.
\newblock An end-to-end spatio-temporal attention model for human action recognition from skeleton data.
\newblock In \emph{Proceedings of the AAAI conference on artificial intelligence}, volume~31, 2017.

\bibitem[Tancik et~al.(2022)Tancik, Casser, Yan, Pradhan, Mildenhall, Srinivasan, Barron, and Kretzschmar]{tancik2022block}
Matthew Tancik, Vincent Casser, Xinchen Yan, Sabeek Pradhan, Ben Mildenhall, Pratul~P Srinivasan, Jonathan~T Barron, and Henrik Kretzschmar.
\newblock Block-nerf: Scalable large scene neural view synthesis.
\newblock In \emph{Proceedings of the IEEE/CVF conference on computer vision and pattern recognition}, pp.\  8248--8258, 2022.

\bibitem[Tang et~al.(2019)Tang, Xu, Sebe, Wang, Corso, and Yan]{tang2019multi}
Hao Tang, Dan Xu, Nicu Sebe, Yanzhi Wang, Jason~J Corso, and Yan Yan.
\newblock Multi-channel attention selection gan with cascaded semantic guidance for cross-view image translation.
\newblock In \emph{Proceedings of the IEEE/CVF conference on computer vision and pattern recognition}, 2019.

\bibitem[Toker et~al.(2021)Toker, Zhou, Maximov, and Leal-Taix{\'e}]{toker2021coming}
Aysim Toker, Qunjie Zhou, Maxim Maximov, and Laura Leal-Taix{\'e}.
\newblock Coming down to earth: Satellite-to-street view synthesis for geo-localization.
\newblock In \emph{Proceedings of the IEEE/CVF conference on computer vision and pattern recognition}, 2021.

\bibitem[Tseng et~al.(2023)Tseng, Li, Kim, Alsisan, Huang, and Kopf]{tseng2023consistent}
Hung-Yu Tseng, Qinbo Li, Changil Kim, Suhib Alsisan, Jia-Bin Huang, and Johannes Kopf.
\newblock Consistent view synthesis with pose-guided diffusion models.
\newblock In \emph{Proceedings of the IEEE/CVF conference on computer vision and pattern recognition}, 2023.

\bibitem[Vaswani et~al.(2017)Vaswani, Shazeer, Parmar, Uszkoreit, Jones, Gomez, Kaiser, and Polosukhin]{vaswani2017attention}
Ashish Vaswani, Noam Shazeer, Niki Parmar, Jakob Uszkoreit, Llion Jones, Aidan~N Gomez, {\L}ukasz Kaiser, and Illia Polosukhin.
\newblock Attention is all you need.
\newblock In \emph{Advances in neural information processing systems}, 2017.

\bibitem[Wang \& Liu(2024)Wang and Liu]{wang2024depthanywhere}
Ning-Hsu Wang and Yu-Lun Liu.
\newblock Depth anywhere: Enhancing 360 monocular depth estimation via perspective distillation and unlabeled data augmentation.
\newblock \emph{Advances in Neural Information Processing Systems}, 37, 2024.

\bibitem[Wang et~al.(2022)Wang, Zhang, Zhang, Ouyang, Chen, Chen, and Wen]{wang2022pretraining}
Tengfei Wang, Ting Zhang, Bo~Zhang, Hao Ouyang, Dong Chen, Qifeng Chen, and Fang Wen.
\newblock Pretraining is all you need for image-to-image translation.
\newblock \emph{arXiv preprint arXiv:2205.12952}, 2022.

\bibitem[Wang et~al.(2018{\natexlab{a}})Wang, Liu, Zhu, Tao, Kautz, and Catanzaro]{wang2018high}
Ting-Chun Wang, Ming-Yu Liu, Jun-Yan Zhu, Andrew Tao, Jan Kautz, and Bryan Catanzaro.
\newblock High-resolution image synthesis and semantic manipulation with conditional gans.
\newblock In \emph{Proceedings of the IEEE/CVF conference on computer vision and pattern recognition}, 2018{\natexlab{a}}.

\bibitem[Wang et~al.(2018{\natexlab{b}})Wang, Girshick, Gupta, and He]{Wang_2018_CVPR}
Xiaolong Wang, Ross Girshick, Abhinav Gupta, and Kaiming He.
\newblock Non-local neural networks.
\newblock In \emph{Proceedings of the IEEE/CVF conference on computer vision and pattern recognition}, 2018{\natexlab{b}}.

\bibitem[Woo et~al.(2018)Woo, Park, Lee, and Kweon]{woo2018cbam}
Sanghyun Woo, Jongchan Park, Joon-Young Lee, and In~So Kweon.
\newblock Cbam: Convolutional block attention module.
\newblock In \emph{European conference on computer vision}, 2018.

\bibitem[Workman et~al.(2015)Workman, Souvenir, and Jacobs]{workman2015wide}
Scott Workman, Richard Souvenir, and Nathan Jacobs.
\newblock Wide-area image geolocalization with aerial reference imagery.
\newblock In \emph{Proceedings of the IEEE/CVF International Conference on Computer Vision (ICCV)}, 2015.

\bibitem[Workman et~al.(2017)Workman, Zhai, Crandall, and Jacobs]{workman2017unified}
Scott Workman, Menghua Zhai, David~J Crandall, and Nathan Jacobs.
\newblock A unified model for near and remote sensing.
\newblock In \emph{Proceedings of the IEEE International Conference on Computer Vision}, pp.\  2688--2697, 2017.

\bibitem[Workman et~al.(2022)Workman, Rafique, Blanton, and Jacobs]{workman2022revisiting}
Scott Workman, M~Usman Rafique, Hunter Blanton, and Nathan Jacobs.
\newblock Revisiting near/remote sensing with geospatial attention.
\newblock In \emph{Proceedings of the IEEE/CVF conference on computer vision and pattern recognition}, 2022.

\bibitem[Wu et~al.(2023{\natexlab{a}})Wu, Liu, Zhao, Bui, Lin, Zhang, and Chang]{wu2023harnessing}
Qiucheng Wu, Yujian Liu, Handong Zhao, Trung Bui, Zhe Lin, Yang Zhang, and Shiyu Chang.
\newblock Harnessing the spatial-temporal attention of diffusion models for high-fidelity text-to-image synthesis.
\newblock In \emph{Proceedings of the IEEE/CVF International Conference on Computer Vision}, pp.\  7766--7776, 2023{\natexlab{a}}.

\bibitem[Wu et~al.(2022)Wu, Tang, Jing, Zhao, Qian, Sebe, and Yan]{wu2022cross}
Songsong Wu, Hao Tang, Xiao-Yuan Jing, Haifeng Zhao, Jianjun Qian, Nicu Sebe, and Yan Yan.
\newblock Cross-view panorama image synthesis.
\newblock \emph{IEEE Transactions on Multimedia}, 2022.

\bibitem[Wu et~al.(2023{\natexlab{b}})Wu, Zheng, and Cham]{wu2023ipoldm}
Tianhao Wu, Chuanxia Zheng, and Tat-Jen Cham.
\newblock Panodiffusion: Depth-aided 360-degree indoor rgb panorama outpainting via latent diffusion model, 2023{\natexlab{b}}.

\bibitem[Xie et~al.(2021)Xie, Wang, Yu, Anandkumar, Alvarez, and Luo]{xie2021segformer}
Enze Xie, Wenhai Wang, Zhiding Yu, Anima Anandkumar, Jose~M Alvarez, and Ping Luo.
\newblock Segformer: Simple and efficient design for semantic segmentation with transformers.
\newblock In \emph{Advances in neural information processing systems}, 2021.

\bibitem[Xie et~al.(2023)Xie, Zhang, Li, Zhang, and Zhang]{xie2023s}
Ziyang Xie, Junge Zhang, Wenye Li, Feihu Zhang, and Li~Zhang.
\newblock S-nerf: Neural radiance fields for street views.
\newblock \emph{arXiv preprint arXiv:2303.00749}, 2023.

\bibitem[Xue et~al.(2023)Xue, Huang, Sun, Song, and Zhang]{xue2023freestyle}
Han Xue, Zhiwu Huang, Qianru Sun, Li~Song, and Wenjun Zhang.
\newblock Freestyle layout-to-image synthesis.
\newblock In \emph{Proceedings of the IEEE/CVF conference on computer vision and pattern recognition}, pp.\  14256--14266, 2023.

\bibitem[Zhai et~al.(2017)Zhai, Bessinger, Workman, and Jacobs]{zhai2017predicting}
Menghua Zhai, Zachary Bessinger, Scott Workman, and Nathan Jacobs.
\newblock Predicting ground-level scene layout from aerial imagery.
\newblock In \emph{Proceedings of the IEEE/CVF conference on computer vision and pattern recognition}, 2017.

\bibitem[Zhang et~al.(2023)Zhang, Rao, and Agrawala]{zhang2023adding}
Lvmin Zhang, Anyi Rao, and Maneesh Agrawala.
\newblock Adding conditional control to text-to-image diffusion models.
\newblock In \emph{Proceedings of the IEEE/CVF international conference on computer vision}, pp.\  3836--3847, 2023.

\bibitem[Zhang et~al.(2018)Zhang, Isola, Efros, Shechtman, and Wang]{zhang2018unreasonable}
Richard Zhang, Phillip Isola, Alexei~A Efros, Eli Shechtman, and Oliver Wang.
\newblock The unreasonable effectiveness of deep features as a perceptual metric.
\newblock In \emph{Proceedings of the IEEE/CVF conference on computer vision and pattern recognition}, 2018.

\bibitem[Zhao et~al.(2023)Zhao, Chen, Chen, Bao, Hao, Yuan, and Wong]{zhao2023uni}
Shihao Zhao, Dongdong Chen, Yen-Chun Chen, Jianmin Bao, Shaozhe Hao, Lu~Yuan, and Kwan-Yee~K Wong.
\newblock Uni-controlnet: All-in-one control to text-to-image diffusion models.
\newblock \emph{Advances in Neural Information Processing Systems}, 36:\penalty0 11127--11150, 2023.

\bibitem[Zheng et~al.(2023)Zheng, Zhou, Li, Qi, Shan, and Li]{zheng2023layoutdiffusion}
Guangcong Zheng, Xianpan Zhou, Xuewei Li, Zhongang Qi, Ying Shan, and Xi~Li.
\newblock Layoutdiffusion: Controllable diffusion model for layout-to-image generation.
\newblock In \emph{Proceedings of the IEEE/CVF conference on computer vision and pattern recognition}, 2023.

\bibitem[Zhou et~al.(2024)Zhou, Lin, Shan, Wang, Sun, and Yang]{zhou2024drivinggaussian}
Xiaoyu Zhou, Zhiwei Lin, Xiaojun Shan, Yongtao Wang, Deqing Sun, and Ming-Hsuan Yang.
\newblock Drivinggaussian: Composite gaussian splatting for surrounding dynamic autonomous driving scenes.
\newblock In \emph{Proceedings of the IEEE/CVF Conference on Computer Vision and Pattern Recognition}, pp.\  21634--21643, 2024.

\bibitem[Zhu et~al.(2017)Zhu, Park, Isola, and Efros]{zhu2017unpaired}
Jun-Yan Zhu, Taesung Park, Phillip Isola, and Alexei~A Efros.
\newblock Unpaired image-to-image translation using cycle-consistent adversarial networks.
\newblock In \emph{Proceedings of the IEEE/CVF International Conference on Computer Vision (ICCV)}, 2017.

\bibitem[Zhu et~al.(2021)Zhu, Yang, and Chen]{Zhu_2021_CVPR}
Sijie Zhu, Taojiannan Yang, and Chen Chen.
\newblock Vigor: Cross-view image geo-localization beyond one-to-one retrieval.
\newblock In \emph{Proceedings of the IEEE/CVF conference on computer vision and pattern recognition}, 2021.

\end{thebibliography}
\bibliographystyle{tmlr}

\newpage
\appendix
\section*{\huge Appendix}
\newcommand{\appendixhead}%
\appendixhead


\section{Dataset Details}
The Brooklyn and Queens dataset that we adopt is an urban dataset with complex scenes. The original dataset contains overhead images downloaded from Bing Maps (zoom level 19) and street-view panoramas from Google Street View. We further collect the satellite images that are center-aligned with the panoramas, to enable a fair comparison with cross-view synthesis methods. Visualizations produced using our approach are based on a GeoDiffusion variant trained on the original dataset (not center-aligned). As the Brooklyn and Queens dataset consists of urban scenes with diverse buildings and transient objects (e.g., cars, pedestrians), it is more challenging for panorama synthesis than commonly used cross-view synthesis datasets like CVUSA~\cite{workman2015wide} and CVACT~\cite{liu2019lending}, which consist of mostly rural scenes. 

\section{Training Details}
Our model is trained under batch size 16 deployed over 4 NVIDIA A100 80GB GPUs.

\section{Details about Geospatial Attention}


During the local-level geospatial attention extraction, we first use a CNN encoder to extract features from the nearby street-level panoramas and the overhead image. The encoder consists of a ResNet-50 pretrained model and a 2D convolution with ReLU activation function. After that max-pooling and average pooling operations are applied along the channel dimensions. The shape of each panorama feature $F_i$ and the overhead image feature $S(l_t)$ are both $H\times W \times 2$.


For the distance information, from the input and target locations $(l_i,l_t)$, we calculate the geometric feature maps of the haversine distance $d$ between $l_i$ and $l_t$. The shape of the output distance feature is \( H \times W\times 1 \); For the orientation $\theta$ from the source to target panorama $l_t$, we compute it by rotating the original pixel rays, which are initially in the east-north-up coordinate frame, by computing the compass bearing between the source and target panorama, so that $[0,1,0]$ points to the target location. The shape of the orientation feature is $H \times W\times3$.

The distance, orientation, the panorama feature map and the satellite image feature map are concatenated to a $H\times W \times 8$ feature tensor. Firstly, the feature tensor is passed through a $3\times3$ and a $5\times5$ convolution layer separately and concatenated to a temporal feature map, then the temporal feature map is passed to a $1\times 1$ convolution layer with softmax activation function to get the spatial attention map $P_{i,t}$. After that the attention map $P_{i,t}$ is passed through an upsample layer to generate the local geospatial attention for the panorama image. The attention mask represents the geometric relationship between the nearby panorama and the target panorama. 

For the global geospatial attention, the extracted local geospatial attention $ P_{1,t},P_{2,t},\cdots,P_{i,t} $ are concatenated, and passed through the average pooling layers, the upsample module, the batch norm layer with sigmoid layer in sequence, and we get the global attention map. In the dataset, street-view near panoramas are collected along the streets, and in practical usage, the target locations are also mainly close to the streets. The global-level geospatial attention excludes the occlusions of buildings and guides the model to focus on the area around the target location with rich semantic information from the overhead view.

\begin{figure*}[t!]
\centering
\includegraphics[width=0.9\linewidth]{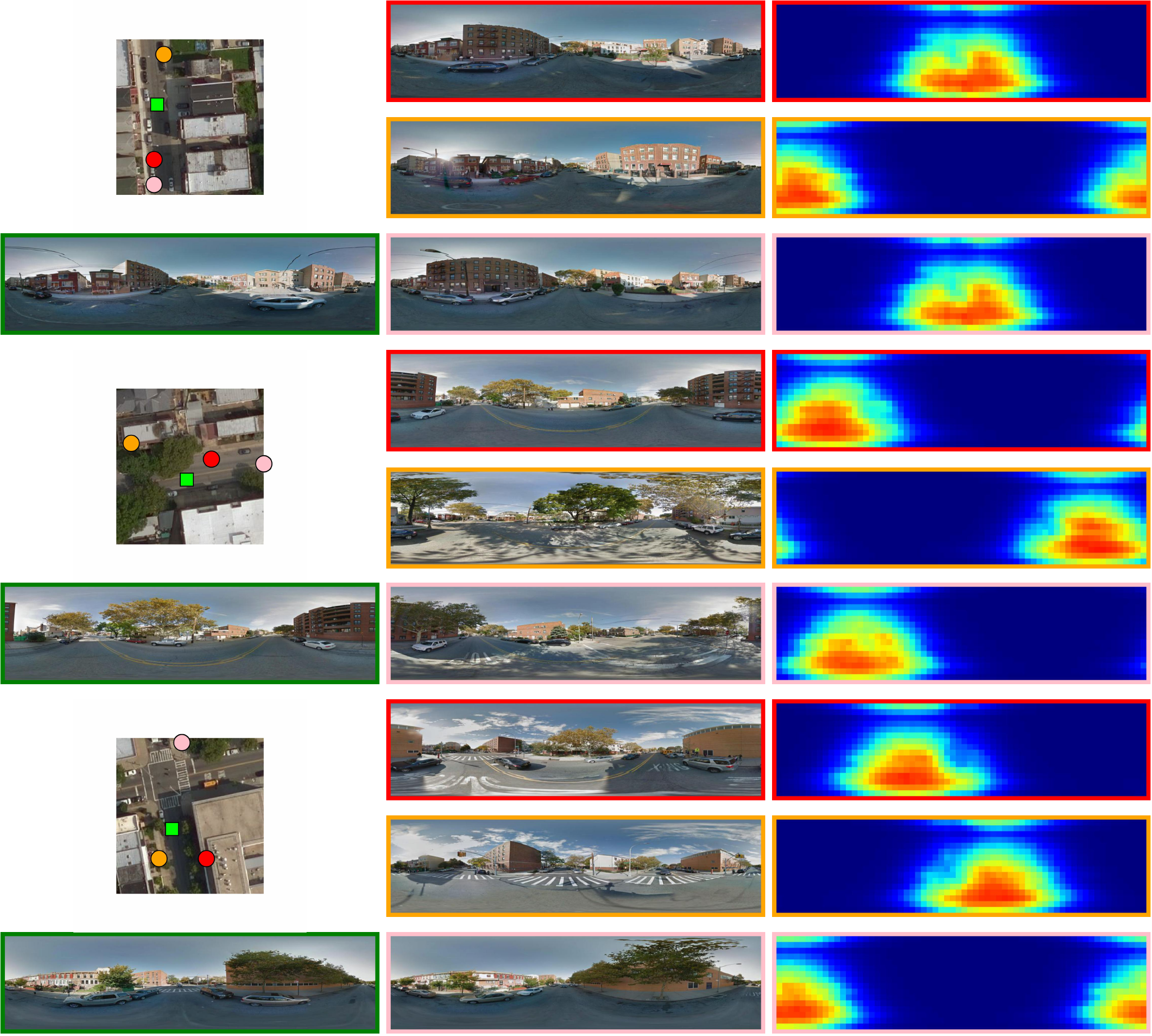}
\caption{Additional visualizations of local geospatial attention. The target location is represented by a green square in the satellite image. The nearby street-level panoramas (color-coded borders) are represented by same-colored circles in the satellite image. }
\label{fig:local_attention}
\end{figure*}

We also show further visualizations of geospatial attention in \figref{local_attention} and \figref{global_att}.

\begin{figure}[ht]
    \centering
    \begin{subfigure}[t]{0.8\linewidth}
        \centering
        \includegraphics[width=\linewidth]{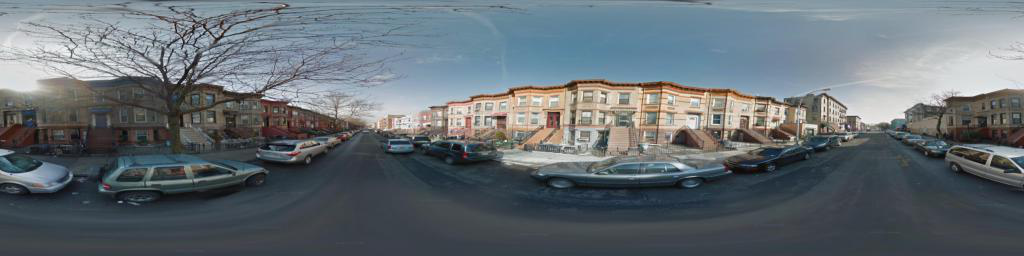}
        \caption{Ground-truth}
        \label{fig:1a}
    \end{subfigure}
    \vspace{1em}
    \begin{subfigure}[t]{0.8\linewidth}
        \centering
        \includegraphics[width=\linewidth]{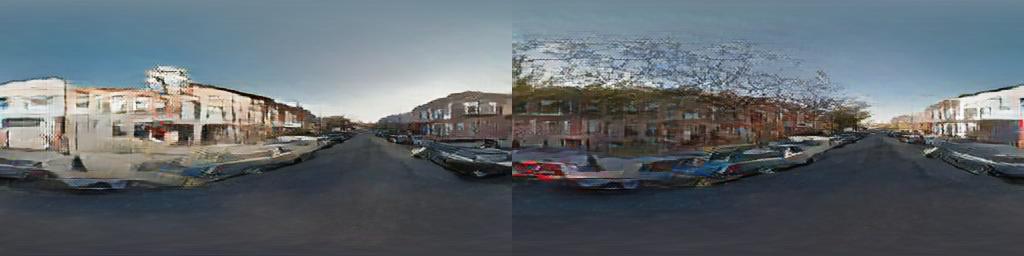}
        \caption{Sat2Density-rotated}
        \label{fig:1b}
    \end{subfigure}
    \vspace{1em}
    \begin{subfigure}[t]{0.8\linewidth}
        \centering
        \includegraphics[width=\linewidth]{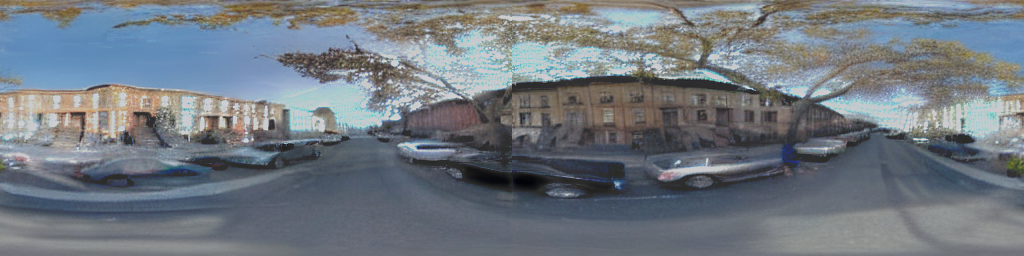}
        \caption{Ours-rotated}
        \label{fig:1c}
    \end{subfigure}
    \caption{Comparison about seamless of generated images.}
    \label{fig:results}
\end{figure}

\section{About Seamlessness}
We follow the evaluation protocol established in prior cross-view synthesis works and evaluate the seamlessness of the generated panoramas, comparing the results versus previous state-of-the-art cross-view synthesis method, Sat2Density~\cite{qian2023sat2density}. We randomly select one result and rotate it $180^{\circ}$ about the vertical direction, which is shown in \figref{results}. The generated panoramas are not absolutely seamless, but compared with Sat2Density, our result is better in the concatenated area. 

Note that different from works about panorama outpainting~\citep{cylin_painting,wu2023ipoldm}, we leverage the geospatial relationships among various input modalities to perform cross-view panorama generation, which existing panorama outpainting methods are not designed to accommodate.

\begin{figure}[t!]
    \centering
    \includegraphics[width=0.8\linewidth]{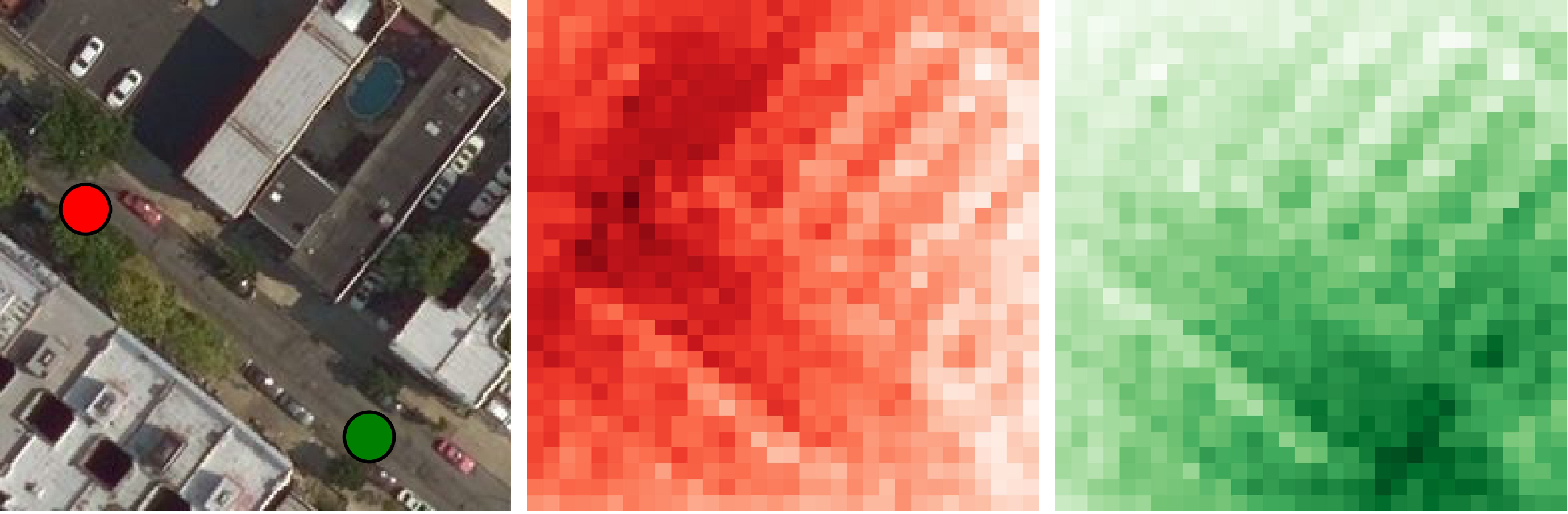}
    \caption{Visualization of global geospatial attention. The color-coded attention maps for two target locations are shown, corresponding to the same-colored dots in the satellite image. Darker colors represent more salient regions.}
    \label{fig:global_att}
\end{figure}

\section{Additional Visualizations on Brooklyn}

\begin{figure*}[t!]
    \centering
    \begin{subfigure}[b]{0.105\linewidth}
        \includegraphics[width=\linewidth]{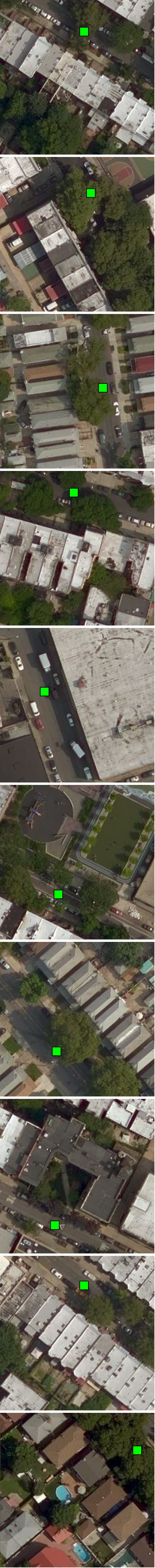}
        \caption{Sat}
        \label{fig:1a}
    \end{subfigure}
    %
    %
    \begin{subfigure}[b]{0.21\linewidth}
        \includegraphics[width=\linewidth]{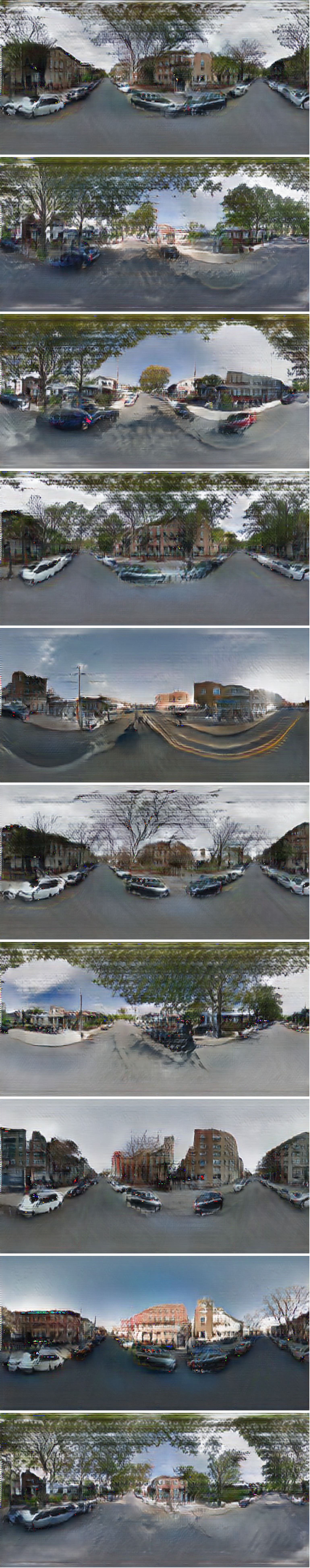}
        \caption{PanoGAN}
        \label{fig:1c}
    \end{subfigure}
    %
    \begin{subfigure}[b]{0.21\linewidth}
        \includegraphics[width=\linewidth]{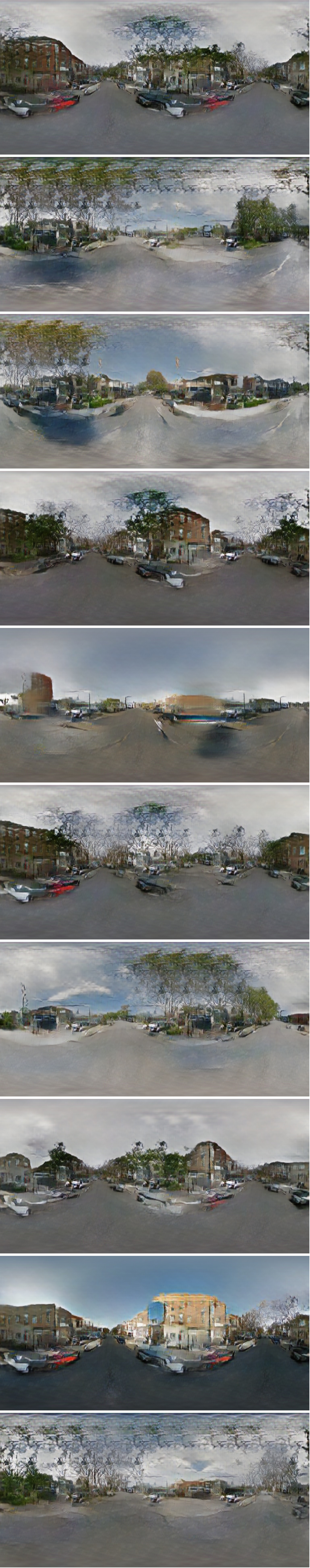}
        \caption{Sat2Density}
        \label{fig:1d}
    \end{subfigure}
    %
    \begin{subfigure}[b]{0.21\linewidth}
        \includegraphics[width=\linewidth]{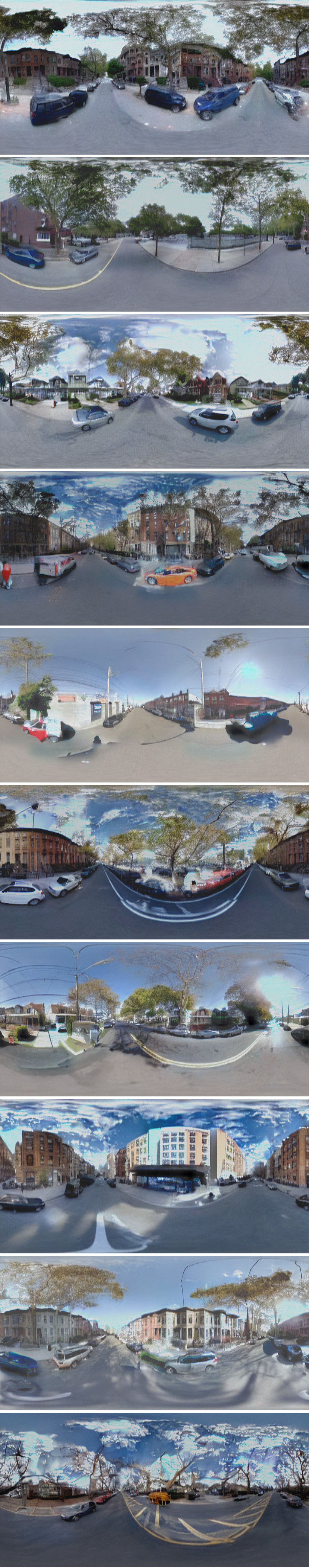}
        \caption{\textbf{Ours}}
        \label{fig:1e}
    \end{subfigure}
    %
    \begin{subfigure}[b]{0.21\linewidth}  
        \includegraphics[width=\linewidth]{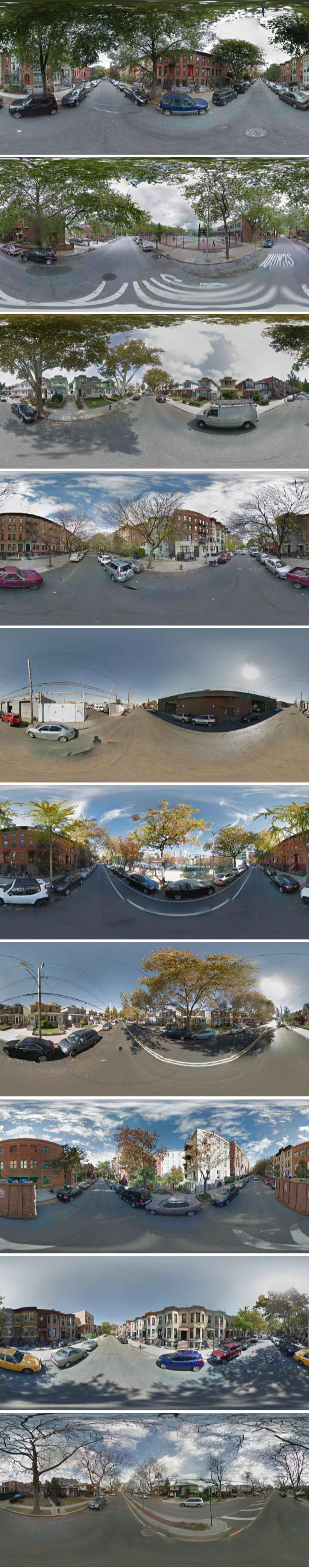}
        \caption{Ground Truth}
        \label{fig:1f}
    \end{subfigure}
    
    \caption{Additional qualitative results versus baselines on the Brooklyn test subset. The target location is represented by a green square in the satellite image. The cross-view synthesis methods that we compare with are trained on our collected center-aligned satellite images. Our approach, which integrates nearby street-level panoramas, is trained on the original satellite images (without center-aligned with the target location). Our method shows better results both geometrically and semantically. }
    \label{fig:results_brooklyn}
\end{figure*}

We show further visualizations on the Brooklyn subset in \figref{results_brooklyn}. Our model is trained on the original unaligned dataset. Compared with the cross-view synthesis methods that are trained on data with center-aligned satellite images, our GeoDiffusion method can synthesize accurate results both semantically and geometrically.

\section{Additional Visualizations on Queens}
\begin{figure*}[t!]
    \centering
    \begin{subfigure}[b]{0.105\linewidth}
        \includegraphics[width=\linewidth]{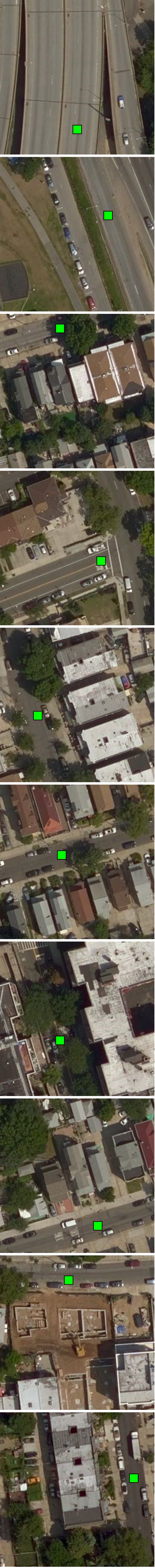}
        \caption{Sat}
        \label{fig:1a}
    \end{subfigure}
    %
    %
    \begin{subfigure}[b]{0.21\linewidth}
        \includegraphics[width=\linewidth]{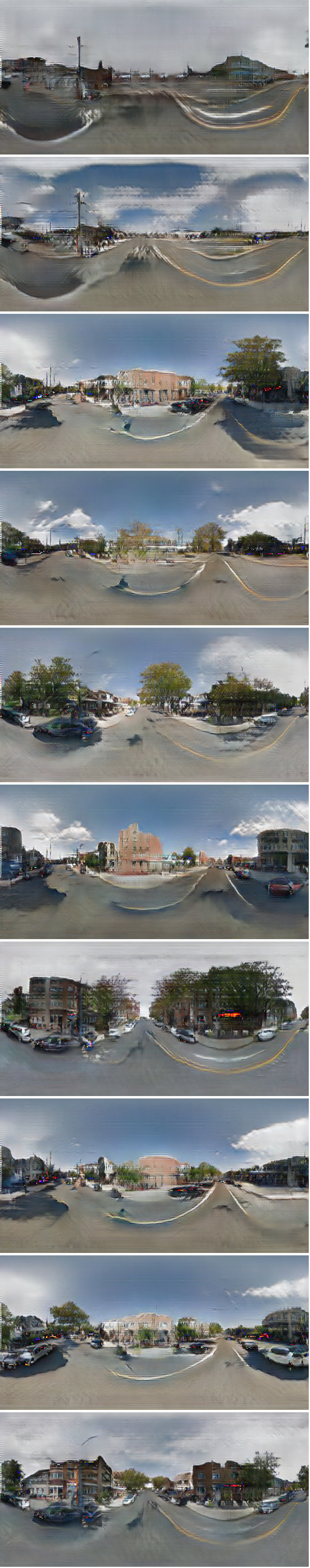}
        \caption{PanoGAN}
        \label{fig:1c}
    \end{subfigure}
    %
    \begin{subfigure}[b]{0.21\linewidth}
        \includegraphics[width=\linewidth]{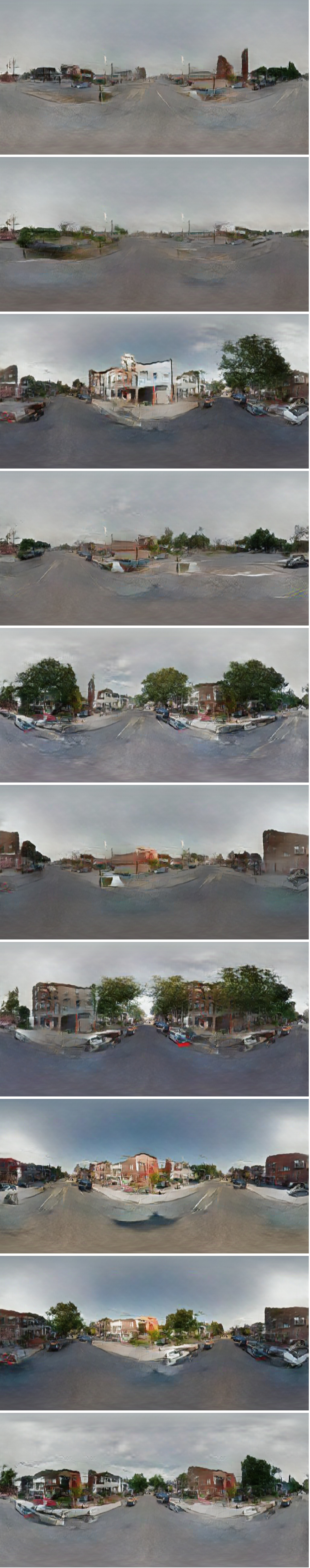}
        \caption{Sat2Density}
        \label{fig:1d}
    \end{subfigure}
    %
    \begin{subfigure}[b]{0.21\linewidth}
        \includegraphics[width=\linewidth]{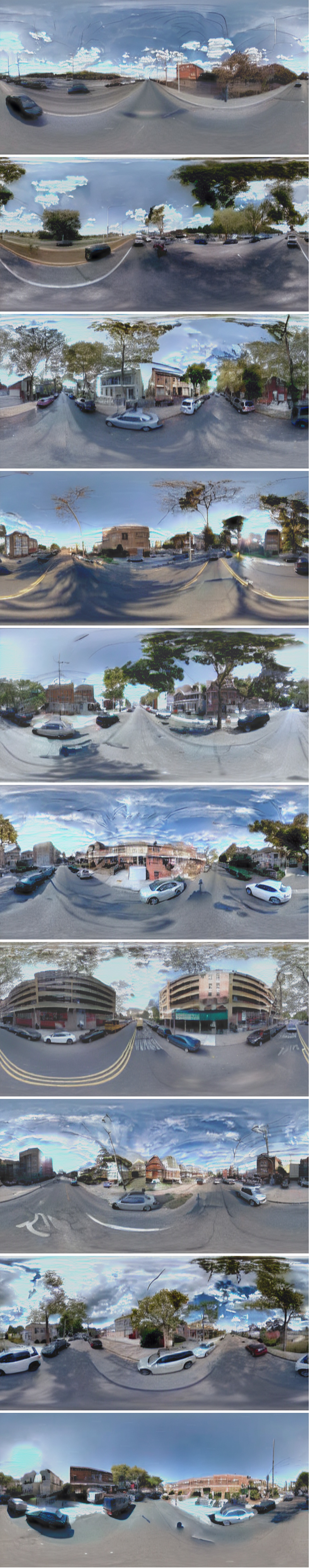}
        \caption{\textbf{Ours}}
        \label{fig:1e}
    \end{subfigure}
    %
    \begin{subfigure}[b]{0.21\linewidth}  
        \includegraphics[width=\linewidth]{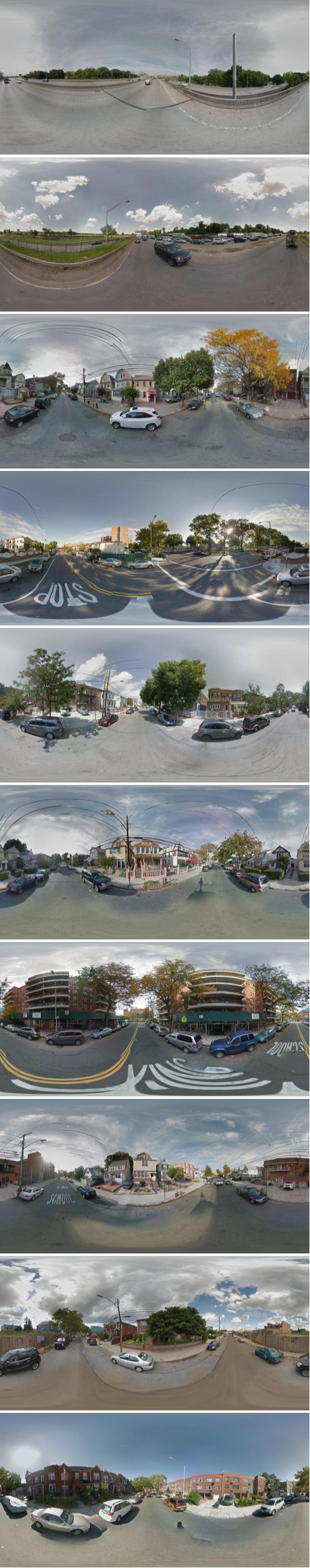}
        \caption{Ground Truth}
        \label{fig:1f}
    \end{subfigure}
    
    \caption{Cross-domain results versus baselines on the Queens subset. The target location is represented by a green square in the satellite image. The cross-view synthesis methods that we compare with are trained on our collected center-aligned satellite images. Our approach, which integrates nearby street-level panoramas, does not rely on the center-aligned satellite image. Our method not only generates more realistic results when compared to baselines, but more accurate results both semantically and geometrically when compared to the ground truth.}
    \label{fig:results_queens}
\end{figure*}

We show further cross-domain visualization results on Queens subset in \figref{results_queens}. Compared with the cross-view synthesis methods that trained on data with center-aligned satellite images, our GeoDiffusion model can get accurate results with geometry accuracy. Results show the generalization ability of our proposed method.

\section{Camera Model}
In the dataset, the satellite images approximate parallel projection and street-view panoramas follow spherical equirectangular projection. The panoramas in the Brooklyn and Queens dataset are with $360^\circ$ horizontal and $180^\circ$ vertical field of view, and are pre-rectified so that the center column line of the panorama represents the north direction.

\section{Discussions}
In practical usage, we expect to collect street-view panoramas and satellite images ahead of time and preload them into the database. Our model is able to synthesize the panorama of a given location by using both the satellite images and the pre-loaded panoramas, creating a dense panorama field from the sparse panoramas. As our method reduce the reliance on aligned data, it is possible to allow synthesizing a bunch of locations in the satellite image region simultaneously, which improve the efficiency of creating dense panorama field.


\end{document}